\documentclass[journal]{IEEEtran}

\usepackage[left=0.625in,right=0.625in,top=0.75in,bottom=1in]{geometry}
\usepackage{ifpdf}
\usepackage{cite}
\usepackage{amsfonts,amssymb}
\usepackage{array}
\usepackage{dblfloatfix}
\usepackage{url}
\usepackage{amsmath}
\usepackage{ragged2e}
\usepackage{graphicx}
\usepackage{verbatim}
\usepackage{float}
\usepackage{lipsum}
\usepackage{multicol, blindtext}
\usepackage{caption}
\usepackage{subcaption}
\usepackage{xcolor}
\usepackage{xurl}
\definecolor{mypink2}{RGB}{0, 0, 255}
\definecolor{green}{RGB}{0, 128, 0}
\usepackage{multirow}
\usepackage{array}
\usepackage{tabularx}
\usepackage[acronym]{glossaries}
\usepackage{algorithm}
\usepackage{algorithmic}
\usepackage{hyperref}
\begin{document}
\title{\fontsize{14pt}{14pt}\selectfont Human-in-the-Loop Machine Learning for Safe and Ethical Autonomous Vehicles: Principles, Challenges, and Opportunities}

\author{Yousef~Emami,~\IEEEmembership{Senior Member,~IEEE,}
         Mohammadhossein~Homaei,~\IEEEmembership{Senior Member,~IEEE,} 
        Miguel~Gutiérrez~Gaitán,~\IEEEmembership{Senior Member,~IEEE,} 
         Luis~Almeida,~\IEEEmembership{Senior Member,~IEEE,} 
         
         Kai~Li,~\IEEEmembership{Senior Member,~IEEE,} 
        Hui~Huang,~\IEEEmembership{Senior Member,~IEEE,} 
        and~Zhu~Han,~\IEEEmembership{Fellow,~IEEE,}

\thanks{Copyright (c) 2026 IEEE. Personal use of this material is permitted. However, permission to use this material for any other purposes must be obtained from the IEEE by sending a request to pubs-permissions@ieee.org.}    
}
\maketitle

\begin{abstract}
Machine Learning (ML) has become central to Autonomous Vehicles (AVs), supporting perception, prediction, planning, control, and decision-making in dynamic environments. However, achieving full autonomy in cluttered and complex scenarios, such as intricate intersections, diverse scenes, varied trajectories, and complex missions, remains challenging; moreover, data labeling is still a major bottleneck. These limitations motivate Human-in-the-Loop Machine Learning (HITL-ML), in which human input is incorporated through validation, annotation, task organization, reward design, action correction, preference feedback, and supervisory intervention. To advance safe and ethical autonomy, this paper presents a tutorial survey of HITL-ML for AVs, focusing on Curriculum Learning (CL), Human-in-the-Loop Reinforcement Learning (HITL-RL), Human-in-the-Loop Large Language Models (HITL-LLMs), Active Learning (AL), and ethical principles. We first review CL methods that structure training from simple to complex tasks, covering navigation, path planning, obstacle avoidance, data collection, landing, intersection handling, motion planning, and UAV swarm coordination. We then examine HITL-RL through reward shaping, action injection, demonstrations, preference-based feedback, and interactive learning, emphasizing improved learning efficiency, safer policy exploration, and real-time intervention. Next, we review HITL-LLM through collaboration and oversight and specify key challenges. After that, we discuss AL for perception, anomaly detection, semantic mapping, object detection, vehicle recognition, and security-related classification. Ethical principles are reviewed as technical requirements for transparency, accountability, human oversight, safety, security, regulatory compliance, and reliability of human input. Finally, we present two case studies on HITL-DQN and HITL-in-context learning for UAV-assisted IoT data collection and identify open research directions.
\end{abstract}

\begin{IEEEkeywords}
 Autonomous Vehicles, Uncrewed Aerial Vehicles,
 Human-In-The-Loop, Machine Learning, Curriculum Learning, Reinforcement Learning, Large Language Model, Active Learning, Ethical Principles 
\end{IEEEkeywords}
\IEEEpeerreviewmaketitle
\begin{table}[htbp]
    \centering
    \caption{List of Acronyms}
    \label{tab:acronyms}
    \begin{tabular}{|l|l|}
        \hline
        \multicolumn{1}{|c|}{\textbf{Acronym}} & \multicolumn{1}{|c|}{\textbf{Definition}} \\
        \hline
        AD & Autonomous Driving \\ \hline
        AL & Active Learning \\ \hline
        AoI & Age of Information \\ \hline
        AUV & Autonomous Underwater Vehicle \\ \hline
        AVs & Autonomous Vehicles \\ \hline
        CL & Curriculum Learning \\ \hline
        CNN & Convolutional Neural Network \\ \hline
        CPSs & Cyber-Physical Systems \\ \hline
        CV & Connected Vehicle \\ \hline
        DDPG & Deep Deterministic Policy Gradients \\ \hline
        DL & Deep Learning \\ \hline
        DQN & Deep-Q-Network \\ \hline
        DRL & Deep Reinforcement Learning \\ \hline
        EVs & Electric Vehicles \\ \hline
        GNN & Graph Neural Networks \\ \hline
        HCI & Human-computer interaction \\ \hline
        HITL & Human-In-The-Loop \\ \hline
        ICL  & In-Context Learning  \\ \hline
        MDP & Markov Decision Process \\ \hline
        ML & Machine Learning \\ \hline
        MLR & Multiple Linear Regression \\ \hline
        PPO & Proximal Policy Optimization \\ \hline
        RAG & Retrieval-Augmented Generation \\ \hline
        RL & Reinforcement Learning \\ \hline
        SVM & Support Vector Machine \\ \hline
        SVR & Support Vector Regression \\ \hline
        TD3 & Twin Delayed Deep Deterministic \\ \hline
        UAV & Uncrewed Aerial Vehicle \\ \hline
        VLM & Vision Language Model \\ \hline
        XGBoost & eXtreme Gradient Boosting \\ \hline
    \end{tabular}
\end{table}
\begin{figure*} [ht] 
       \centering
       \captionsetup{justification=raggedright}
        \includegraphics[scale=0.43]{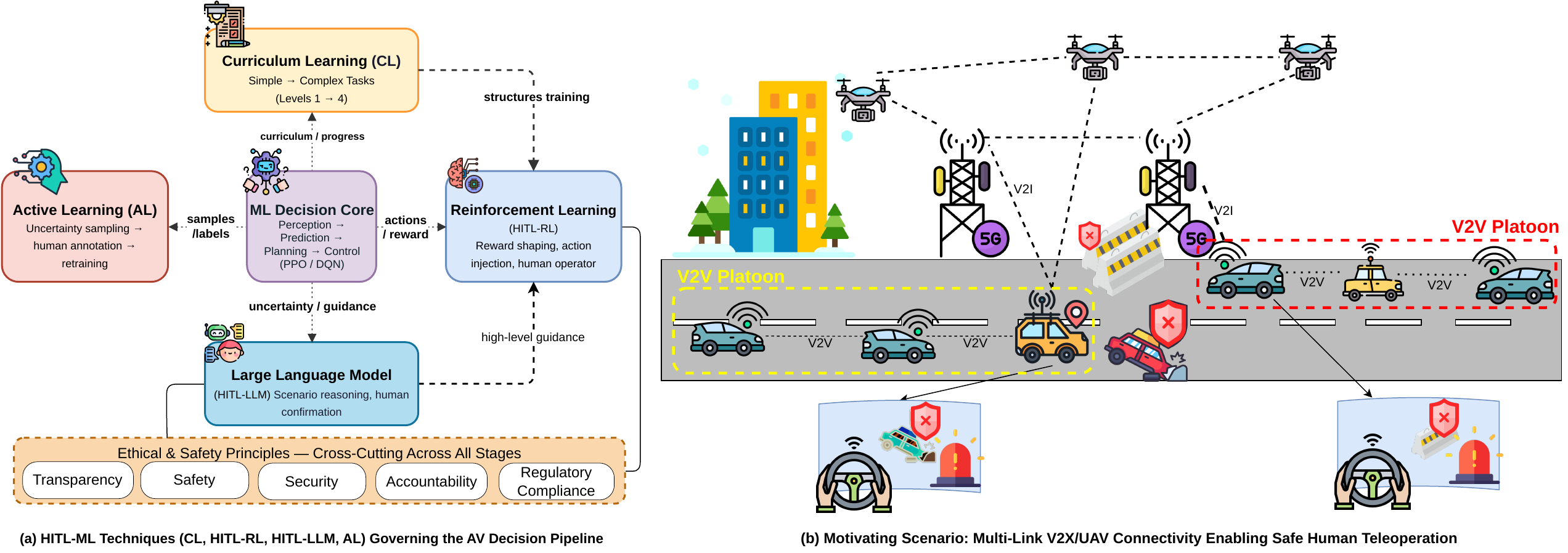}
    	\caption{ The workflow of HITL-ML framework for autonomous vehicle decision-making. The framework integrates CL, HITL-RL, HITL-LLM, and AL to enhance the perception–prediction–planning–control pipeline while ensuring transparency, safety, security, accountability, and regulatory compliance.}
    	\label{fig1}

\end{figure*}

\section{Introduction}
Recent developments in Autonomous Vehicles (AVs) have improved efficiency and safety \cite{7823109}. AVs have become a major driver of automation, connectivity, and intelligent mobility. The global AV market size was valued at USD 1,500.3 billion in 2022 and is projected to grow from USD 1,921.1 billion in 2023 to USD 13,632.4 billion by 2030 \cite{fortune}. Commercially envisioned and available self-driving cars can perform functions such as lane changing, highway driving, and autonomous parking by sensing the environment using built-in technologies \cite{8809568}. Ground-based autonomous systems can be paired with aerial systems to support surveillance, monitoring, traffic management, remote sensing, and autonomous control \cite{9356608}\cite{lisinska2021autonomous}. Current trends in AVs include leveraging Machine Learning (ML) exploring Large Language Models (LLMs) for high-level reasoning and human--machine interaction\cite{11361285}, tightening security \cite{8753541}\cite{8762090}, and facilitating data and command dissemination \cite{8844775}\cite{8422933} to reduce accidents, fatalities, injuries, pollution, and traffic congestion while fostering a more equitable transportation infrastructure.

ML is central to AV operation because it enables data-driven perception, prediction, planning, and control from large-scale sensory and contextual data.
ML algorithms facilitate the detection and classification of objects, pedestrians, and other vehicles, as well as the prediction of their future trajectories. These predictive insights are critical for enabling AVs to make real-time braking, turning, acceleration, and steering decisions. The use of AVs therefore relies on ML-driven systems to perform key functions such as perception\cite{9684905}, state estimation\cite{10262351}, path planning\cite{9696360}, traffic rule compliance, and decision making in dynamic and uncertain environments. However, these functions become more difficult to validate when AVs face rare events, ambiguous scenes, incomplete data, or safety-critical decisions that require human supervision.
\par
\subsection{Motivation and Background}
Human-in-the-loop Machine Learning (HITL-ML)  
integrates human expertise into ML pipelines through supervision, feedback, intervention, and validation. Human intuition and judgment help AVs handle situations that are not well represented in training data~\cite{twosigma2024}. Decision-making for ethical dilemmas in unavoidable accidents requires explicit treatment of context, operational constraints, responsibility, and acceptable risk, which are difficult to capture using learned policies alone.

Human input and guidance can enhance ML models by improving data quality, correcting unsafe actions, shaping rewards, calibrating uncertain decisions, and supporting accountability in safety-critical operation.
\par
Humans can be incorporated into AV learning pipeline through collaboration and oversight. In collaboration, humans and ML models provide complementary capabilities. Humans excel at capturing region-specific behaviors and cultural norms, providing context and insights that ML algorithms might miss. ML models can trigger human assistance when confidence is low, novelty is high, or the expected risk exceeds a predefined threshold.
\par
Imagine an AV driving on a local road where pedestrians frequently cross unpredictably and outside designated crosswalks. In such a scenario, the AV’s ML model may struggle to adapt, resulting in unsafe or overly conservative actions such as abrupt stops or failure to yield appropriately. When the AV encounters these challenges, it can trigger human intervention to prevent unsafe behaviors and refine its responses. A human operator familiar with the local conditions can override or adjust the model’s actions, supplementing the reward system to prioritize pedestrian safety, even in situations that deviate from the AV’s original training data. This human input helps ensure that the AV can navigate the environment safely and effectively despite the limitations of its original training.

\begin{table*}[t]
\centering
\caption{Comparison of Existing Surveys and Tutorials Related to HITL-ML for Autonomous Vehicles}
\label{tab:comparison_surveys}
\renewcommand{\arraystretch}{1.2}
\resizebox{\textwidth}{!}{%
\begin{tabular}{lcccccccccc}
\hline
\textbf{Reference} &
\textbf{HITL} &
\textbf{CL} &
\textbf{HITL-RL} &
\textbf{AL} &
\textbf{LLMs} &
\textbf{Ethics} &
\textbf{AVs} &
\textbf{Tutorial} &
\textbf{Case Studies} &
\textbf{Main Focus} \\
\hline

~\cite{7911263} &
\checkmark & -- & -- & -- & -- & -- & -- & -- & -- &
Human factors in mobile networks \\

~\cite{mosqueira2023human} &
\checkmark & $\sim$ & $\sim$ & $\sim$ & -- & -- & -- & -- & -- &
General HITL-ML techniques \\

~\cite{wu2022survey} &
\checkmark & -- & -- & \checkmark & -- & -- & -- & -- & -- &
Human interaction for data quality \\

~\cite{10.1007/978-3-030-83906-2_20} &
\checkmark & -- & $\sim$ & -- & -- & $\sim$ & -- & -- & -- &
Human-machine teaming \\

~\cite{10530996} &
\checkmark & -- & $\sim$ & $\sim$ & -- & $\sim$ & -- & -- & -- &
HITL methodologies and challenges \\

~\cite{WILCHEK2023106376} &
\checkmark & -- & -- & $\sim$ & -- & -- & $\sim$ & -- & -- &
HITL for computer vision \\

~\cite{10.5555/3455716.3455897} &
-- & \checkmark & $\sim$ & -- & -- & -- & -- & -- & -- &
Curriculum Learning for RL \\

~\cite{SOORI202354} &
-- & $\sim$ & -- & -- & -- & -- & $\sim$ & -- & -- &
AI/ML for robotics \\

~\cite{9016391} &
-- & $\sim$ & -- & -- & -- & $\sim$ & \checkmark & -- & -- &
AI in autonomous vehicles \\

~\cite{7029083} &
\checkmark & -- & -- & -- & -- & -- & $\sim$ & -- & -- &
HITL Cyber-Physical Systems \\

~\cite{10258330} &
-- & -- & -- & -- & -- & -- & \checkmark & -- & -- &
End-to-end autonomous driving \\

~\cite{ZHAO2024122836} &
-- & -- & -- & -- & -- & $\sim$ & \checkmark & -- & -- &
Autonomous driving technologies \\

~\cite{9044647} &
-- & -- & -- & -- & -- & \checkmark & \checkmark & -- & -- &
Ethics in AVs \\

~\cite{9304618} &
-- & -- & -- & -- & -- & \checkmark & \checkmark & -- & -- &
Ethical decision-making for AVs \\

~\cite{11361285} &
-- & -- & -- & -- & \checkmark & -- & \checkmark & -- & -- &
LLMs for AV testing \\

~\cite{11264491} &
-- & -- & -- & -- & \checkmark & $\sim$ & \checkmark & -- & -- &
LLMs/VLMs for AVs \\

\hline
\textbf{This Tutorial} &
\textbf{\checkmark} &
\textbf{\checkmark} &
\textbf{\checkmark} &
\textbf{\checkmark} &
\textbf{\checkmark} &
\textbf{\checkmark} &
\textbf{\checkmark} &
\textbf{\checkmark} &
\textbf{\checkmark} &
\textbf{Unified HITL-ML framework for safe and ethical AVs} \\
\hline
\end{tabular}}
\end{table*}
ML models are effective in tasks such as predictive analytics and risk assessment, but they still require monitoring to ensure safety and alignment with operational objectives (oversight). Tesla’s Autopilot, for instance, may face limitations when driving on local roads in the presence of people and relies on the driver's supervision to navigate effectively \cite{tesla2024modelsmanual}. Google’s self-driving cars still have also been reported to make mistakes, such as being confused by fixed-gear cyclists at stop signs \cite{washingtonpost}. In October 2023, Cruise’s permit to test AVs in California was suspended after a serious accident in which a car dragged a woman across the road. Waymo also recalled some of its vehicles after two collided with the same tow truck within minutes \cite{cnn2024}. This ongoing need for human involvement underscores that the timeline for widespread consumer deployment remains uncertain\cite{Litman2015AutonomousVI}. These examples illustrate that the effective deployment of AVs requires a hybrid approach where human oversight continues to play a crucial role in ensuring safety and reliability.

Ethical principles must be embedded in AVs to align their behavior with operational, legal, and societal requirements. To this end, ethical theories can be translated into constraints, objectives, and oversight mechanisms within the decision-making processes of AVs. However, ethical decision-making in AVs faces several challenges. One major challenge is the complexity of ethical dilemmas that require difficult trade-offs. 
for example, in urban driving scenarios where the system must reduce collision risk while balancing the safety of passengers, pedestrians, and other road users. Another challenge is the need to adapt ethical decision-making processes in real time. Ensuring transparency and accountability in these processes is also crucial. HITL-ML can improve self-assessment and scene understanding in AVs to assist in managing complex situations such as accidents and supporting ethically informed decisions in real-time. 
\par
As shown in Fig. \ref{fig1}, the workflow of HITL-ML for AVs comprises several key steps\cite{wu2022survey}: 1) developing an ML model for prediction, decision-making, or control, 2) applying human validation and annotation with expert oversight, 3) conducting model training where input data is organized and refined, and 4) real-world deployment while considering ethical implications. Learning algorithms such as Proximal Policy Optimization (PPO) \cite{zhao2024real} can be used within this pipeline, while human experts monitor the learning process, override unsafe actions in critical situations, and modify or supplement the reward function. Moreover, Active Learning (AL) can identify uncertain or informative instances and refer them to human annotators~\cite{8961120}. The model is trained after human validation, annotation, and feedback from human experts. Meanwhile, Curriculum Learning (CL) informed by human expertise structures training data or tasks for AV ~\cite{automata}, facilitating and expediting the learning process. Finally, the model is deployed in the real world, considering ethical implications.
\par
Rapid progress in user interface design~\cite{8057581}, regulatory frameworks~\cite{9803769}, AV security~\cite{10.1145/3365996}\cite{10592471}, and explainable AI~\cite{9583190} has increased the feasibility of deploying HITL approaches for AVs. Improved user interfaces provide intuitive controls and real-time feedback, while regulatory requirements promote human oversight for safety and ethical compliance. Developments in AV security help mitigate cyber threats, and research in explainable AI enhances the transparency of decision-making processes, making it easier for human operators to understand and interact with AVs. Together, these advancements enable more effective human oversight and intervention, crucial for the safe and ethical operation of AVs.

\par
\subsection{Existing Surveys and Tutorials}
In the literature, several works consider HITL-ML. For example, Duan etal.~\cite{7911263} provide a survey on understanding and exploiting the human factor in different contexts of mobile networks. Expanding on HITL-ML strategies, Mosqueira-Rey etal.~\cite{mosqueira2023human} present a comprehensive review of state-of-the-art techniques, highlighting different strategies for incorporating humans into the ML process. From a data perspective, Wu etal.~\cite{wu2022survey} survey HITL-ML, addressing how human interaction can enhance data quality and model performance.  Rajendran etal.~\cite{10.1007/978-3-030-83906-2_20} review the main concepts and methods for HITL learning, paving the way for frameworks that facilitate human-machine teaming through safe learning and anomaly prediction.  Methodologies, challenges, and opportunities associated with real-world applications of HITL systems are reviewed by Kumar et al.~\cite{10530996}, who discuss key issues such as data quality, bias, and user engagement. Wilchek etal.~\cite{WILCHEK2023106376} review the literature on HITL, emphasizing the role of human interaction in enhancing computer vision model performance and accuracy. However, these surveys are mostly general-purpose or application-specific and do not provide a tutorial treatment of HITL-ML for safe and ethical AVs across CL, HITL-RL, AL, and ethical decision-making.

\begin{figure} [htb] 
       \centering
        \captionsetup{justification=raggedright}
        \includegraphics[ scale=0.95 ]{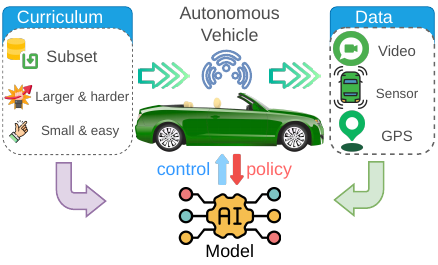}
    	\caption{With CL for AVs tasks are divided into smaller, manageable parts, progressing from easy to more difficult challenges. Data associated with each part is sequentially fed to the model to enhance learning efficiency and improve overall performance. }
    	\label{fig3}
\end{figure}    

Specifically, in the realm of CL within RL, Narvekar etal.~\cite{10.5555/3455716.3455897} provide a framework and an overview of existing CL methods. They classify these methods based on their assumptions, capabilities, and goals, offering insights into how CL can systematically improve RL performance. Soori etal.~\cite{SOORI202354} examine recent advances in AI, ML, and DL in advanced robotic systems, and discuss their applications in robot modeling, control, and automation. Ma etal.~\cite{9016391} review the use of AI in AV development, and elaborate on critical issues such as functional  
requirements and challenges in implementing AI-driven solutions for AVs. From a computer vision system design perspective, Nunes etal.~\cite{7029083} present a review of HITL-Cyber-Physical Systems (CPSs), providing insights into how human factors can be integrated into CPSs to enhance their functionality and reliability. AD technologies have seen substantial advancements, driven by developments in ML and AI. Chib and Singh~\cite{10258330} review the end-to-end AD stack, categorizing automated driving tasks using neural networks in an end-to-end manner. By surveying essential components and cutting-edge technologies, Zhao etal.~\cite{ZHAO2024122836} explore the development and implementation of AD systems. Their survey highlights the key technologies that enable AD and the challenges associated with their deployment. Both Wang etal.~\cite{9044647} and Moura etal.~\cite{9304618} address ethical dilemmas associated with different aspects of AVs. Nevertheless, these works treat CL, AI-based AV development, HITL-CPSs, AD architectures, and AV ethics largely as separate topics, leaving room for a tutorial survey that connects HITL-ML mechanisms with safe and ethical AV operation.

Recently, the use of LLMs for AVs has attracted attention. Zhao etal. \cite{11361285} provide a systematic classification of LLM functions across the distinct phases of scenario-based testing within autonomous driving. Tian etal. \cite{11264491} provide a comprehensive review of existing research on the application of LLMs and Vision Language Models (VLMs) to AVs, with a particular focus on modular integration, end-to-end integration, data generation, evaluation platforms, datasets, and benchmarks.However, these works focus on foundation-model integration and testing rather than HITL-ML mechanisms for safe and ethical AV operation. Table \ref{tab:comparison_surveys} compares existing surveys.

\subsection{Contribution}
In this paper, we review HITL-ML for safe and ethical AV operation, with emphasis on how human input can improve learning, supervision, intervention, and decision-making in complex scenarios, including busy local roads, intricate intersections, varied trajectories, and mission-oriented operation. Our paper differs from others in the literature by emphasizing the critical role of HITL-ML in supporting the safe and ethical operation of AVs. We focus on CL, HITL-RL, HITL-LLM-based methods, and AL (see Table~\ref{tab2}, for a comparative summary of recent research papers on these techniques) and then we examine ethical principles tailored to the unique challenges and opportunities of AVs. 

The key contributions of this paper are listed as follows:

\begin{itemize}

\item We present the first tutorial survey on HITL-ML techniques applied to AVs convering ML pipelines based on CL, RL, LLMs and AL, thus providing a consistent view of this research topic, currently fragmented along each specific technique. This survey leverages the integrated view to highlight synergies between different techniques. 

\item We provide a comprehensive overview of the application of CL in AVs to improve generalization and convergence speed by structuring the training process from simpler to more complex tasks. We highlight the synergy between CL and RL and how their combination can reduce training time and improve AV performance on tasks such as navigation, path planning, and obstacle avoidance.
     
    \item We provide an in-depth overview of the latest HITL-RL methods and their applications, highlighting how human involvement can enhance the RL process through techniques such as reward shaping, action injection, and interactive learning. We also review examples from recent studies showing various HITL-RL frameworks that utilize human feedback to refine AV learning processes and %ensure
   support real-time safety interventions.We present a HITL Deep Q‑Network (DQN) framework for UAV‑assisted IoT data collection, where a rule‑based oracle selectively overrides the agent’s node selection when it is both uncertain (low Q‑value gap) and ignoring a critically aged node.
 
    \item We provide an overview of HITL-LLM-based methods for AVs, highlighting how HITL frameworks integrate human judgment into LLM, allowing operators to intervene when uncertainty or risk is high. We further present a second case study on oracle-assisted in-context learning, in which an edge-based LLM proposes UAV scheduling and velocity decisions that are validated against a deterministic safety oracle enforcing hard battery, channel-quality, and velocity constraints, illustrating how HITL principles extend from RL-based to LLM-based decision-making in UAV resource allocation.

    \item We review the literature on AL, focusing on how AL optimizes the annotation process, accelerates AV development, and enhances system robustness by leveraging human expertise. We also explore how integrating AL with ML techniques such as DL, SVMs, and symmetry-based analysis enhances the efficiency and accuracy of tasks %like
   such as anomaly detection, semantic mapping, object detection, and vehicle recognition in AVs.

    \item We investigate the ethical considerations of HITL-ML for AVs, highlighting the integration of HITL-ML to enhance self-recognition, scene recognition, and real-time ethical decision-making. We also address ethical concerns such as accountability and differential privacy, and examine implementable mechanisms and performance metrics.

\end{itemize}

\subsection{Paper Structure and Organization}

\begin{figure*}[!t]
    \centering
    \includegraphics[width=\textwidth]{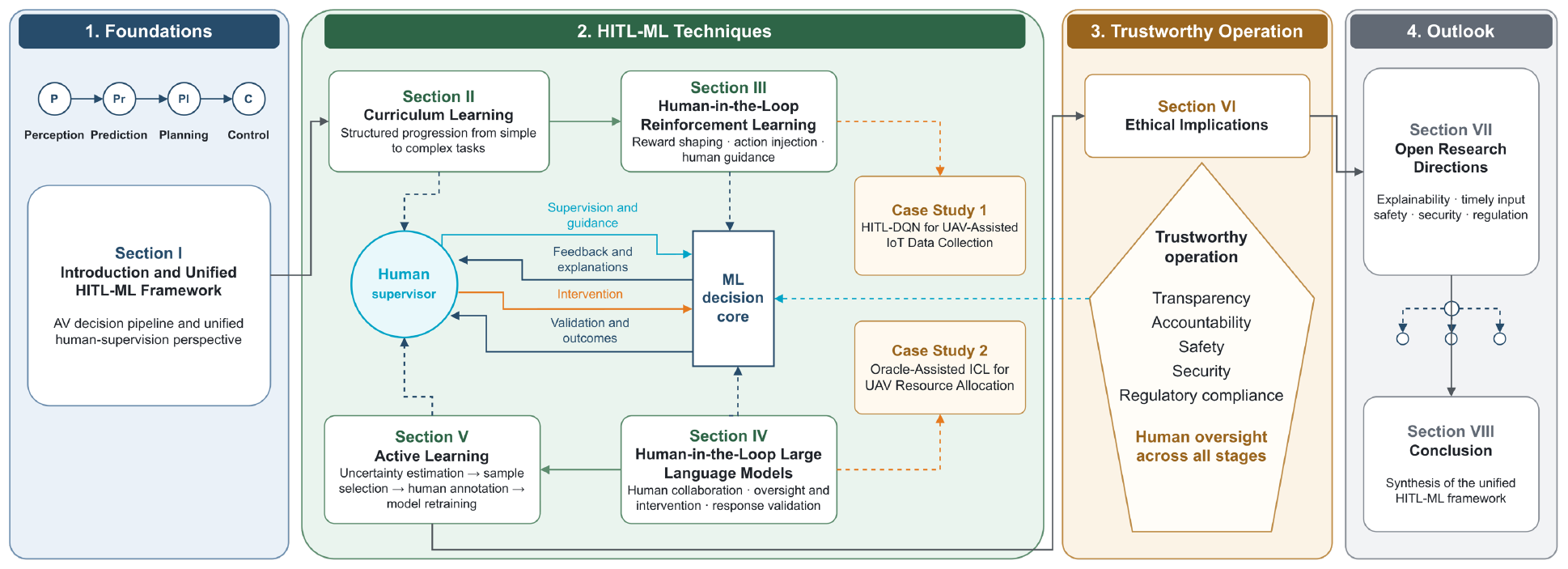}
    %\caption{Overall structure and organization of the paper.}
\caption{Organization of the tutorial, showing the progression from the foundations of HITL-ML to curriculum learning, HITL reinforcement learning, HITL large language models, and active learning. The figure also highlights the human--ML supervision loop, the two case studies, trustworthy operation, open research directions, and conclusion.}
    \label{fig:paper_structure}
\end{figure*}

Fig.~\ref{fig:paper_structure} presents the overall organization and structure of this paper.

% \begin{figure*}[!t]
%     \centering
%     \includegraphics[width=18cm,height=8cm]{hand.png}
%     \caption{Overall structure and organization of the paper.}
%     \label{fig:paper_structure}
% \end{figure*}
% Section~\ref{sec2} introduces CL and its significant role in improving learning speed in AVs. 
% Section~\ref{sec3} introduces HITL-RL and how humans can cooperate in the RL process in AVs. 
% Section~\ref{sec4} introduces HITL-LLMs and how humans can improve the reliability of LLMs' responses.
% Section~\ref{sec5} discusses AL and examines its role in facilitating data annotation in AVs. 
% Section~\ref{sec6} addresses the ethical aspects of AVs and provides guidelines to achieve ethical AVs using HITL-ML. 
% Section~\ref{sec7} highlights open research directions and, finally, we conclude the paper in Section~\ref{sec8}. Table~\ref{tab345} presents a list of the acronyms used throughout this paper for easy reference.

\begin{table*}[ht]
\caption{Comparative Summary of Recent Research Papers Exploring HITL-ML.}
\label{tab2}
\centering
\begin{tabularx}{\textwidth}{|p{0.7cm}|X|p{1.5cm}|p{4cm}|}
\hline
 \multicolumn{1}{|c|}{\textbf{Ref}} & \multicolumn{1}{|c|}{\textbf{Objective}} & \multicolumn{1}{|c|}{\textbf{ML Technique}} & \multicolumn{1}{|c|}{\textbf{Scenario}} \\
\hline
\cite{li2023uav} & assist the UAV in obstacle avoidance. & DRL & Navigation / UAV \\
\cite{WU202375} & improve the agent’s learning ability and performance. & DRL & Intelligence / AV \\
\cite{8976170} & tackle sparse reward and low learning efficiency. & RL & Exploration / AUV \\
\cite{drones7050311} & facilitate mapless obstacle avoidance. & DRL & Navigation / UAV swarm \\
\cite{9793564} & improve the learning performance of RL & DRL & AD \\
\cite{9261709} & mitigate in-flight UAV anomalies by using semi-supervised SVM & AL & Anomaly detection / UAV \\
\cite{ruckin2023informative} & develop a new path planning framework for AL & AL & Path planning / Semantic mapping \\
\cite{zhang2022unknown} & detect unknown network attacks in the UAV network. & AL & IDS / UAV \\
\cite{8961120} & facilitate manual labeling of large-scale surveillance data. & AL & VTR / ITS \\
\cite{10051636} & learn the decentralized flocking and obstacle avoidance policy & CL & Obstacle avoidance / UAV swarm \\
\cite{8500603} & reduce the required training time and improve agent performance & CL & AD \\
\cite{s21051766} & enable mission planning for controlling a UAV & ML & Mission planning / UAV \\
\cite{liang2022exploring} & address 3D object detection in AV & AL & Object detection / AV\\
\cite{YAN2023108091} & study collision-free flocking control problem for scalable and large-scale fixed-wing UAV & CL & Flocking problem / UAV swarm \\
\cite{9782734} & learn an end-to-end urban driving policy for CARLA driving simulator & CL & Driving simulator \\
\cite{11264491} & review of the current research on LLM and VLM applications in AVs & VLM/LLM & AVs\\
\cite{11361285} & categorizing the roles played by LLMs across various phases of scenario-based testing & LLM & Automated Driving Systems\\
\hline
\end{tabularx}
\end{table*}

\section{Curriculum Learning} \label{sec2}

CL is an ML technique that structures the training process by starting with simple tasks and gradually progressing to more complex ones. This approach enhances generalization and convergence speed, making it useful for training AVs~\cite{park2021indoor}. Recent studies have applied CL to AV-related settings in different ways, including curriculum-based multi-agent DRL for decentralized flocking and obstacle avoidance, collision-free flocking control in large-scale UAV swarms, and adaptive curriculum strategies for efficient target search in disaster scenarios using UAV swarms\cite{7192644}. These applications illustrate how CL autonomous agents through progressively challenging tasks to improve learning efficiency and task performance.

\begin{table}[h!]
\caption{Key Techniques for Navigation and Data Collection}
\label{tab10}
\centering
\scriptsize
\begin{tabular}{|p{0.5cm}|p{1.75cm}|p{1.75cm}|p{3cm}|}
\hline
\textbf{Ref} & \textbf{Technique} & \textbf{Application} & \textbf{Strengths and Limitations} \\
\hline
\cite{10051636} & Curriculum-based Multi-agent DRL & Flocking and obstacle avoidance for UAVs & Learns decentralized policies; requires knowledge transfer mechanisms. \\
\hline
\cite{YAN2023108091}  & Curriculum Learning with Multi-agent DRL & Collision-free flocking for large UAV swarms & Scalable to large swarms; gradually increases swarm size. \\
\hline
\cite{10156524} & Curriculum-based DRL & UAV navigation in complex 3D environments & Improves navigation reliability; relies on human-defined curriculum. \\
\hline
\cite{SEONG2024124379} & MARL with Curriculum Learning and Evolutionary Strategy & Data collection from wireless sensor nodes with UAVs & Promotes MARL scalability; integrates energy-efficient charging mechanisms; enhances AoI minimization and energetic sustainability. \\
\hline
\cite{drones7110676} & Automatic Curriculum Learning with TD3 & Collaborative autonomous landing of UAV on UGV & Offers rapid localization and pose estimation; handles different UGV motions and wind interference; breaks down tasks into simpler subtasks. \\
\hline
\end{tabular}
\label{tab:flocking_navigation}
\end{table}

Fig. \ref{fig3} illustrates the concept of CL for AVs. Tasks are divided into smaller, manageable parts, progressing from simple to more difficult challenges. This method allows AVs to build foundational skills with simpler tasks before gradually tackling more complex scenarios, thereby improving learning efficiency and performance over time. The structured progression helps the vehicle handle increasingly difficult situations with greater accuracy and reliability. In the realm of AVs, the combination of CL and RL can reduce the required training time, improve agent performance, and help learn better policies \cite{8500603}. AVs require mastery of complex tasks such as navigation, path planning, and obstacle avoidance. CL can aid the learning process by ordering these tasks from simple to complex.

\subsection{Literature Review}
\subsubsection{Navigation and Data Collection}
Curriculum learning can improve collision avoidance in UAV swarms \cite{10246260} by gradually increasing the complexity of the tasks that the UAVs have to solve during training. This approach ensures that the UAVs first master simpler tasks, such as avoiding static objects or other UAVs in a sparse environment, before moving on to more challenging scenarios, such as dynamic environments with multiple moving objects or denser UAV formations. A particular challenge in UAV swarms is collision avoidance, where the UAVs must avoid other UAVs as well as fixed and moving objects. Yan etal. \cite{10051636} study the flocking and obstacle avoidance problem for multiple fixed-wing UAVs in cluttered environments. A curriculum-based multi-agent DRL method is developed to learn
decentralized flocking and obstacle avoidance policies. In this context, the collision-avoiding flocking task is decomposed into multiple subtasks, and the number of subtasks to be solved is gradually increased. In this approach, the authors introduce a multi-agent actor-critic DRL algorithm and two knowledge transfer mechanisms for online learning and offline transfer, respectively. Similarly, Yan etal. \cite{YAN2023108091} study collision-free flocking control 
for scalable and large-scale fixed-wing UAV swarms by integrating CL with multi-agent DRL. The problem considers both collision avoidance and the large population of UAVs in the swarm. The key idea is to divide the training process into several learning phases and gradually expand the scale of UAV swarms. In addition, Braathen etal. \cite{10156524} investigate UAV navigation in complex 3D environments and develop a solution using curriculum-based DRL. The human expert divides the training into ten levels with an increasing range of possible numbers of obstacles in the environment. The leveraged curriculum exposes the UAV to simpler environments initially and progressively to more complex ones.

he focus of unmanned aerial vehicle (UAV) path planning includes challenging tasks such as obstacle avoidance and efficient target reaching in complex environments. Building upon these fundamental challenges, an additional need exists for agents that can handle diverse missions like round-trip navigation without requiring retraining for each specific task. In our study, we present a path planning method using reinforcement learning (RL) for a fully controllable UAV agent. We combine goal-conditioned RL and curriculum learning to enable agents to progressively master increasingly complex missions, from single-target reaching to round-trip navigation.

\begin{table}[h!]
\caption{Key Techniques for AD and UAV Swarm Navigation}
\label{tab20}
\centering
\scriptsize
\begin{tabular}{|p{0.5cm}|p{1.75cm}|p{1.75cm}|p{3cm}|}
\hline
\textbf{Ref} & \textbf{Technique} & \textbf{Application} & \textbf{Strengths and Limitations} \\
\hline
\cite{8500603} & CL with DRL & AD at urban intersections & Reduces training time; improves performance; uses automated curriculum creation; avoids human handcrafting. \\
\hline
\cite{xiao2022collaborative} & Adaptive Embedded Curriculum & UAV swarm target search in disaster management & Adjusts task difficulty based on success rate; handles sparse rewards. \\
\hline
\cite{9782734} & PPO with Curriculum Learning & Urban driving policy for CARLA simulator & Guides agent to better policy; improves learning efficiency. \\
\hline
\cite{automata} & RD-ACPPO & Self-driving tasks at unsignalized intersections & Adaptive curriculum learning; allocates importance weights dynamically; demonstrates effective curriculum transition timings and high task success rate. \\
\hline
\cite{auto} & Multi-agent scenario-based framework & CARLA simulator for AD & Practical for multi-agent learning; generates auto-curricula from scenario descriptions. \\
\hline
\cite{10422140} & Curriculum Learning with GNN-RL & Motion planning in AD & Incorporates map and traffic rule information; decomposes learning task into sub-tasks with increasing difficulty; includes rewards for temporal-logic-based traffic rules. \\
\hline
\end{tabular}
\label{tab:autonomous_driving_uav_swarm}
\end{table}

Path planning in cooperative vehicular platooning remains a challenging task due to the need for efficient navigation, obstacle avoidance, and target reaching in complex and dynamic environments. Beyond basic navigation, following the platoon leader states, making continual decisions in keeping proper distance to its front member, platoon members are often required to perform diverse missions, such as prediction for non-platoon vehicles, being supportive in managing a cut-in operation (specially in the case of long platoons) without requiring retraining for each individual task. To address this challenge, RL combined with CL can enable Connected and Autonomous Vehicles (CAV), as potential platoon members, to progressively learn increasingly complex missions, starting from simple single-target reaching tasks and advancing toward more complex navigation objectives. For example, starting from a short homogeneous platoon~\cite{11174279} and advancing the training process toward a long heterogeneous platoon~\cite{10136271}. However, dynamic environments introduce additional difficulties due to the presence of moving obstacles, changing weather conditions, and surrounding traffic volume which require platoons to adapt their behavior in real time. Traditional DRL approaches often face limitations in these scenarios, including slow convergence and the need for large amounts of training data. Therefore, efficient learning strategies that improve adaptability and reduce training requirements are essential for reliable platoon navigation.
\par
Moreover, data collection~\cite{9174950} and UAV landing tasks require structured task and training-data organization. Seong etal. \cite{SEONG2024124379} address the collection of data from wireless sensor nodes to minimize the Age of Information during data collection, while considering the energy sustainability of UAVs. They introduce Multi-Agent Reinforcement Learning (MARL) in combination with CL and an evolutionary strategy to improve the performance of traditional MARL setups as the number of agents increases. Their method promotes MARL scalability and integrates energy-efficient charging mechanisms that 
improve system performance in large-scale deployments. Moreover, the collaborative autonomous landing of a quadrotor UAV on a moving Unmanned Ground Vehicle (UGV) is challenging due to the need for accurate real-time tracking of the UGV and the adjustment of the landing policy. Wang et al.~\cite{drones7110676} propose the Landing Vision System to offer rapid localization and pose estimation of the UGV. Then, they design an Automatic CL approach in conjunction with the Twin Delayed Deterministic Policy Gradient (TD3) that breaks down UAV landing control into a series of simpler subtasks, and learns the landing tasks under different conditions of UGV motions and wind interference. Table \ref{tab10} summarizes the above findings. Traditional MARL setups face with performance issue when scaling up the number of agents. Curriculum learning facilitates the management of complex UAV swarm tasks, such as collision avoidance, flocking, and navigation, by gradually increasing the complexity of training scenarios. By applying curriculum learning, MARL scalability is improved but also effectively enhancing system performance in large-scale deployments. CL combined with multi-agent DRL improves learning efficiency, scalability, and the ability of UAVs to achieve safe and coordinated flight in complex environments.

Collectively, these studies reveal that CL for UAV navigation and data collection is staged along several distinct axes rather than a single generalizable dimension. \cite{10051636} and \cite{YAN2023108091} stage \emph{swarm scale}, incrementally expanding the number of cooperating fixed-wing UAVs while the flocking and obstacle-avoidance task structure remains fixed; \cite{10156524} instead stages \emph{obstacle density} for a single UAV navigating a static 3D environment, with a human expert manually defining ten difficulty levels; \cite{SEONG2024124379} stages \emph{agent count jointly with an energy-sustainability objective}, coupling CL with an evolutionary strategy rather than a purely hand-crafted schedule; and \cite{drones7110676} departs from environmental or population complexity altogether, applying CL to decompose a single control task -- autonomous landing -- into sequential subtasks under varying disturbance conditions such as UGV motion and wind interference. Despite this diversity in staging axis, a common limitation persists across all five works: curricula are \emph{human-defined}, with the number of stages, their pacing, and the difficulty metric fixed a priori rather than learned or adapted online during training. Moreover, only \cite{SEONG2024124379} and \cite{drones7110676} integrate CL with a secondary optimization objective (energy sustainability and pose-estimation accuracy, respectively), whereas \cite{10051636}, \cite{YAN2023108091}, and \cite{10156524} treat CL purely as a mechanism for improving policy convergence and collision-avoidance performance. This suggests that, while CL is well-established for reducing training complexity in UAV swarm and navigation tasks, its integration with auxiliary system-level objectives remains comparatively underexplored.

\subsubsection{Autonomous Driving}

Self-driving tasks such as intersection management and motion planning benefit from CL because CL can structure complex driving problems into progressively harder training stages. Peng et al. \cite{automata} present a reward-driven automated curriculum proximal policy optimization (RD-ACPPO) approach for self-driving tasks at unsignalized intersections. An adaptive CL technique is presented to accommodate progressively more difficult learning tasks in AD. Their approach allows importance weights to be allocated across different curricula during different RL training periods. The RL policy trained by the proposed framework achieved a high task success rate. In addition, Peiss etal. \cite{10422140} develop a CL scheme for motion planning in AD that incorporates traffic rule and map information into a GNN-RL agent whose state is extended by map and traffic rule information. They decompose the complicated learning task by applying CL to the GNN-RL agent. The curriculum's sequence of sub-tasks gradually increases the level of difficulty of the motion planning problem for the agent. Each sub-tasks contains a set of rewards, including rewards for temporal-logic-based traffic rules for speed, safety distance, and braking. Moreover, the authors of \cite{8500603} investigate the challenge of learning AD behaviors at urban intersections with DRL. They combine CL with DRL to reduce the required training time and improve agent performance. They define a set of tasks and use an automated method to create the curriculum for the training process, reducing the burden of human handcrafting. Table \ref{tab20} summarizes the above findings. Automatic curriculum learning improves the training and generalization of reinforcement learning-based autonomous driving agents by dynamically adjusting scenario complexity according to the agent’s learning progress. Unlike fixed training scenarios or manually designed curricula, the approach uses an automated teacher to generate and modify scenarios based on their learning potential, reducing expert bias and improving training efficiency. This enables agents to develop more robust policies that can better handle diverse real-world driving conditions.

CL can facilitate learning and training in the CARLA \cite{dosovitskiy2017carla} simulator and target search for UAV swarms. Anzalone etal. \cite{9782734} aim to learn an end-to-end urban driving policy for the CARLA driving simulator by combining the PPO algorithm with CL. In their approach, the RL phase is divided into several stages of increasing difficulty, so that the agent is guided to learn a better driving policy. The curriculum-based agent is evaluated using different metrics, cities, weather conditions, and traffic scenarios. Furthermore, Brunnbauer etal. \cite{auto} develop a multi-agent, scenario-based AD training and evaluation framework for the CARLA simulator. The results show that the proposed approach is practical for multi-agent learning and can generate auto-curricula from scenario descriptions. In a similar vein, Xiao etal. \cite{xiao2022collaborative} use a UAV swarm for target search in disaster management scenarios. An adaptive embedded curriculum can adjust the difficulty of the task based on the success rate achieved in training and is used to enable UAVs to explore and navigate cluttered 3D environments with sparse rewards more efficiently. Curriculum learning enhances training efficiency in simulated autonomous driving like CARLA simulator, where CL combined with RL enables agents to learn driving policies through staged training and supports evaluation across diverse traffic, weather, and environmental conditions. For UAV swarm target search, adaptive CL adjusts task difficulty based on training performance, helping UAVs learn effective exploration and navigation strategies in complex 3D environments with sparse rewards.

\begin{figure} [ht] 
       \centering
        \captionsetup{justification=raggedright}
        \includegraphics[ scale=0.8]{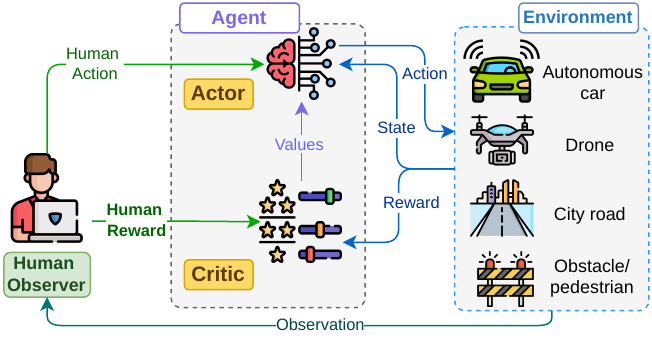}
    	\caption{The actor-critic setup of HITL-RL for AVs allows human experts to inject their actions and guide the agent in critical situations, helping the AV navigate safely and effectively despite unexpected challenges. Additionally, the environmental reward is supplemented with human-defined rewards to guide the learning process. Both the environmental reward and the human reward are fed to the critic.}
    	\label{fig2}
\end{figure} 

\subsection{Key Findings and Insights} 

From the analysis in the previous subsection, the review studies indicate that CL, by systematically increasing the complexity of tasks, not only supports the AV learning process but also reduces training time and improves agent performance. This structured approach is crucial for preparing the vehicle to handle complex scenarios, where HITL-RL can further fine-tune the vehicle's decision-making abilities.
The following open issues emerge from the literature survey:

\subsubsection{
Integration with Transfer Learning}
When developing a curriculum that includes multiple tasks, the intermediate tasks may have different state/action spaces, reward functions, or transition functions compared to the final task. Consequently, transfer learning is essential to extract and pass on reusable knowledge from one task to the next. Traditionally, transfer learning focuses on how to transfer knowledge from one or more source tasks directly to a target task. However, CL extends this concept by considering training sessions in which the agent repeatedly transfers knowledge from one task to another, resulting in a series of end tasks~\cite{10.5555/3455716.3455897} . The integration of transfer learning and CL has practical relevance in AVs. For instance, an AV might first learn basic navigation in a simple environment and transfer the learned knowledge to complex tasks such as obstacle avoidance or object manipulation. It might also be useful in platooning applications to share the navigation knowledge or maneuver coordination knowledge with other platoon members.

\subsubsection{CL vs Hard Example Mining}
When developing a curriculum for AVs, a common approach is to order tasks from easy to hard, which is central to CL. However, this raises the question of whether this incremental approach is always the most effective strategy. An alternative or complementary approach, Hard Example Mining (HEM)~\cite{shrivastava2016training}, shifts the focus from a structured progression to selecting the most challenging examples for training. HEM operates under the assumption that harder examples, which are often identified by the model's loss or gradient magnitude, provide more informative learning opportunities.

\begin{table}[h!]
\caption{Key HITL-DRL Techniques with Reward Shaping}
\label{tab30}
\centering
\scriptsize
\begin{tabular}{|p{0.5cm}|p{1.5cm}|p{1.75cm}|p{3cm}|}
\hline
\textbf{Ref} & \textbf{Technique} & \textbf{Application} & \textbf{Strengths and Limitations} \\
\hline
\cite{li2023uav} & HITL-DRL & UAV navigation & Improves obstacle avoidance and navigation efficiency; requires multiple reward functions. \\
\hline
\cite{10309494} & PbRL & UAV navigation & Uses human feedback for reward; dependent on accurate feedback. \\
\hline
\cite{10541090} & RL with human driving data & AV navigation & Learns reward function from human driving data; complexity in modeling human decisions. \\
\hline
\cite{10334023} & HITL-RL with TD3 & AV lane-change decisions & Encodes human experience in rewards and constraints; guided by human experts to limit unsafe exploration. \\
\hline
\end{tabular}
\label{tab:hitl_drl}
\end{table}

\begin{table}[h!]
\caption{Key HITL-DRL Techniques with Action Injection}
\label{tab40}
\centering
\scriptsize
\begin{tabular}{|p{0.5cm}|p{1.75cm}|p{1.75cm}|p{3cm}|}
\hline
\textbf{Ref} & \textbf{Technique} & \textbf{Application} & \textbf{Strengths and Limitations} \\
\hline
\cite{WU202375} & HITL-DRL with Human Guidance & Autonomous Vehicle navigation & Improves learning ability and safety; complexity in integrating human actions with DRL policy updates. \\
\hline
\cite{10450270} & HITL-DRL Framework & UAV swarm navigation in large-scale 3D urban environments & Allows human intervention to correct dangerous actions; increased complexity in training Actor-Critic networks with human feedback. \\
\hline
 \cite{10381656}  & HITL-DDPG with HCI Feedback & Autonomous UAV navigation in challenging environments & Handling of unforeseen delays due to human input processing. \\
\hline
\cite{HUANG2024100127} & Enhanced HITL-RL for Driving Policy Learning & AD in mixed-traffic environments & Improves safety and traffic flow efficiency; requires effective human supervision, intervention, and demonstration. \\
\hline
\end{tabular}
\end{table}
\subsubsection{Human Defined CL vs Human Data}
This leads to an important consideration: should we always rely on explicitly designed curricula, or can we leverage human data to generate curricula 
in a data-driven manner?. Using real-world driving data from humans allows AVs to be exposed to a broad spectrum of realistic scenarios and behaviors, thereby enhancing their learning process. This data-driven approach may offer a more nuanced and practical training experience than a pre-defined curriculum.

\subsubsection{Co-learning: a multi-agent approach to CL}
Additionally, another possibility is the use of co-learning, a multi-agent approach to CL where the curriculum emerges from the interactions among multiple AVs within the same environment. In this setup, the curriculum evolves adaptively as AVs interact, creating an implicit learning process. This interaction-driven method can support continuous improvement, as the agents learn from each other and adapt to increasingly complex driving tasks over time, as discussed by \cite{10.5555/3455716.3455897} . This dynamic approach may offer advantages by reflecting real-world driving challenges and enabling more robust learning outcomes.

\section{Human-In-The-Loop Reinforcement Learning} 
\label{sec3}

HITL-RL can enhance the RL process by leveraging techniques such as reward shaping, action injection, and interactive learning. These methods assist AVs in tasks such as lane-changing, autonomous driving, obstacle avoidance, navigation, and overall performance enhancement, thereby improving safety and supporting more reliable operation. The setup of HITL-RL for AVs is shown in Fig. \ref{fig2}. In this setup, human experts play a crucial role in monitoring the RL process. They can intervene by injecting actions when necessary and can enhance the system by providing supplementary rewards, known as human rewards. This integration helps keep the learning process safe, reliable, and aligned with human feedback and operational constraints.

\subsection{Literature review}
\subsubsection{Reward Shaping}
In reward shaping, the environmental reward is supplemented with human-defined rewards to guide learning. HITL can switch between multiple UAV reward functions in different situations and assist in constructing interactive reward functions. More generally, human feedback can reduce the need for fully hand-crafted reward functions by providing preferences, corrections, or task-specific guidance during training.

Li etal. \cite{li2023uav} develop a HITL-DRL framework for UAV navigation in large-scale 3D complex environments. This work aims to enable the UAV to avoid obstacles automatically during flight. The role of HITL is to assist the UAV in obstacle avoidance by dynamically switching between multiple UAV reward functions in different situations. The combined approach improves the success rate and navigation efficiency. In a similar vein, Karumanchi etal. \cite{10309494} address the problem of autonomous UAV navigation and develop a practical solution based on Preference-based Reinforcement Learning (PbRL). In PbRL, the manual construction of the reward function is replaced with human feedback. A human is presented with a pair of trajectories followed by the RL agent and is asked to indicate a preference for one over the other. The PbRL algorithm then computes a policy or action sequence using only the set of preferences collected over different trajectories. 
\par

In the AD context, reward functions can be learned from human driving data and encoded with human prior experience. Zhuang etal. \cite{10541090} address the challenges of designing reward functions for complex RL objectives. The framework comprises trajectory sampling, offline preference learning, and RL. Feasible trajectories are generated by sampling target states from a reachable state space; then, a transformer network is trained to model human driving decisions and derive a reward function by comparing generated trajectories with human driving trajectories. Finally, the derived reward function is incorporated into the RL framework to obtain the driving decision policy. In another related effort, Huang etal. \cite{10334023} aim to facilitate mandatory lane-change decisions for AVs in complex environments. In this context, they develop a HITL-RL method based on Twin Delayed Deep Deterministic policy gradient (TD3) for lane changing decisions. They encode human prior experience into the reward function and safety constraints in an offline manner. Also, human experts guide the agents online to limit unsafe exploration during the training process and provide demonstrations in complex scenarios. Table \ref{tab30} summarizes the above findings. In autonomous driving, human experience is incorporated into reward functions and safety constraints, enabling RL agents to learn safer and more effective decision-making policies in complex environments. In UAV navigation, HITL assists in dynamically selecting reward functions and guiding obstacle avoidance, improving navigation success and efficiency. Preference-based reinforcement learning reduces reliance on manually designed rewards by learning from human trajectory preferences.

\subsubsection{Action Injection}
The human can inject actions and guide the agent in critical situations, helping AVs navigate safely and effectively despite unexpected challenges. Also, human guidance is used in real-time to improve the agent's learning ability and performance.

Introducing human guidance into the training loop of DRL algorithms enables AVs to better handle different situations. Further, human intervention can improve UAV swarm safety. 
Wu etal. \cite{WU202375} state that the limited intelligence and capabilities of DRL-based strategies in AV applications prevent them from handling various situations and replacing human drivers. In the proposed HITL-DRL, human guidance is used in real-time to improve the agent's learning ability and performance. Specifically, the human participant observes the agent's real-time training process and assists when necessary. The provided action is prioritized as the output action of the agent to interact with the environment and replaces the original action of the DRL policy. The actor and critic are updated according to the human guidance and experience of the DRL agent. The actor-network can learn from both human guidance through imitation learning and the experience of the interactions through RL. The critic network can evaluate the values of the agent's actions as well as the actions provided by human guidance. Similarly, Li etal. \cite{10450270} develop a HITL-DRL framework for UAV swarm navigation in a large-scale 3D complex urban environment. This framework enables human intervention in the flight of the UAV swarm when necessary to correct unsafe actions. Additionally, the Actor-Critic structure of DDPG has been modified to identify and learn from positive human experiences.
 
Human feedback can enhance UAV obstacle avoidance performance by having the human act as a mentor to the autonomous agent. Lee etal. \cite{10381656} investigate autonomous navigation in challenging environments with the capability to avoid obstacles. The UAV navigates toward its target using the DDPG model, which can handle unexpected situations. HITL feedback is integrated with the DDPG model to enhance obstacle avoidance performance. This feedback is crucial when the pre-trained DDPG model encounters environments to which it is not fully adapted and gives the system the flexibility to handle unforeseen scenarios and make real-time adjustments. For the HCI feedback, the first-person shooter concept is employed, where feedback is provided using keyboard directional keys. Through this HCI-based HITL feedback, real-time control of the UAV flight movement is achieved. 

\begin{table}[h!]
\caption{Key HITL Techniques with Interactive Learning}
\label{tab50}
\centering
\scriptsize
\begin{tabular}{|p{0.5cm}|p{1.75cm}|p{1.75cm}|p{3cm}|}
\hline
\textbf{Ref} & \textbf{Technique} & \textbf{Application} & \textbf{Strengths and Limitations} \\
\hline
\cite{8976170} & Deep Interactive RL & AUV path tracking & Addresses sparse rewards and low learning efficiency; practical implementation challenges. \\
\hline
\cite{drones7050311} & Hybrid Collision Avoidance & UAV collision-free navigation & Effective in complex environments; requires integration of multiple modules. \\
\hline
\cite{9793564} & Human Guidance-based RL & AD & Improves learning efficiency with human guidance; complexity in modeling human behavior. \\
\hline
\cite{s21051766} & ML-based Human-UAV Interaction & UAV mission planning & Facilitates sequential and fast operation; dependent on accurate signal processing. \\
\hline
\cite{agrawal2020model} & Meta-Model for Human-UAV Interaction & Emergency response & Useful in time-critical missions; complexity in specifying interactions and interventions. \\
\hline
\cite{10507015} & Value-based DRL with Human Guidance & AD & Improves training efficiency and adaptability; utilizes human guidance for high-level decision-making in multi-objective lane-change problems. \\
\hline
\cite{10250993} & HITL-RL with Human-Guided Learning & UGV navigation task & Improves RL capabilities with human interventions; includes human-guided objectives, experience replay, and reward shaping. \\
\hline
\end{tabular}
\end{table}
Huang etal. \cite{HUANG2024100127} aim to improve the safety of AD technology, %ensuring 
while maintaining traffic flow efficiency in mixed-traffic environments. They develop an enhanced HITL-RL method for driving policy learning, where the human serves as a mentor to the AI agent, supervising, intervening, and demonstrating in its learning process, and is integrated with DRL to guarantee the safety of the AVs while also improving AV safety while optimizing the traffic flow efficiency. Table \ref{tab40} summarizes the above findings. In AV applications, human actions can replace unsafe agent decisions during training, enabling RL models to learn from both human guidance and environmental interactions. For UAV swarms, human intervention helps correct unsafe navigation behaviors in complex environments and enhances obstacle avoidance by providing real-time adjustments. Overall, action injection improves agent adaptability, safety, and decision-making in unexpected scenarios.

\subsubsection{Interactive Learning}
The agent and human trainer can interact and cooperate to solve tasks. For example, humans can supplement environmental reward and guide learning objectives to improve learning efficiency and RL performance. Zhang etal. \cite{8976170} aim to explore the oceans using autonomous underwater vehicles (AUV). RL can improve their autonomy but still faces practical problems due to sparse rewards and low learning efficiency. They propose a deep interactive RL method for AUV path tracking by combining the advantages of DRL and interactive RL. Since the human trainer cannot give human rewards to the AUV in the operational phase and the AUV needs to adapt to a changing environment when it is in the ocean, a DRL method that simultaneously learns from human rewards and environmental rewards is proposed. Wu etal.  \cite{10250993} improve RL performance in simulation and real-world settings for the navigation task of UGVs using HITL-RL. The framework 
incorporates human intervention through a series of mechanisms to provide demonstrations and improve RL's capabilities. The mechanisms comprise human-guided learning objectives, prioritized human experience replay, and human intervention-based reward shaping. 

Human behavior modeling mechanisms can alleviate the workload on humans, further using meta-model provides a language to describe the roles of UAVs and humans and autonomous decisions. 
Wu etal. \cite{9793564} aim to improve the learning performance of RL by incorporating human guidance. In this regard, they develop a human guidance-based RL framework along with a mechanism for leveraging the experience of human guidance. A TD3-based algorithm was developed to improve algorithmic capabilities in the context of HITL-RL. The performance evaluation was conducted using AD tasks and showed the ability of the developed algorithm to improve learning efficiency compared to state-of-the-art human guidance-based RL. In addition, a human behavior modeling mechanism is proposed to reduce the burden on humans. Additionally, Agrawal etal. \cite{agrawal2020model} investigate semi-autonomous UAVs to support emergency response scenarios. Humans and multiple UAVs must work together as a team to successfully execute a time-critical mission. Therefore, a meta-model is proposed to describe the interaction between humans and multiple UAVs. The meta-model provides a language to describe the roles of UAVs and humans, as well as autonomous decisions. They use two different scenarios involving multiple semi-autonomous UAVs in emergency operations to demonstrate the use of the meta-model by formally specifying human interactions and points of intervention.
\par

HITL-DRL can facilitate UAV operation in complex environments and enable mission planning, further HITL-DRL address high-level decision-making problems in AD. Zhang etal. \cite{drones7050311} state that collision-free navigation for UAVs can facilitate data collection from IoT devices. However, traditional path planning methods are energy and computationally intensive, while DRL-based approaches tend to trap UAVs in complex environments. They propose a hybrid collision avoidance method for real-time UAV navigation in complex environments with unpredictable obstacles. They first develop a HITL-DRL training module for mapless obstacle avoidance and then set up a global planning module that generates waypoints for guidance. In addition, they developed an algorithm for target updating that combines the HITL-DRL training module and the global planning module. In a similar vein, Müezinoğlu et al. \cite{s21051766} develop an intelligent human–UAV interaction approach in real time based on ML using wearable gloves. The objective is to enable mission planning for controlling a UAV. ML-based methods process the signal from the gloves and classify appropriate commands to include in the human–UAV interaction process via the interface. These commands are then incorporated into a task scheduling algorithm to support sequential and fast operation.

Moreover, the authors of \cite{10507015} aim to improve the training efficiency and adaptability of DRL methods to untrained cases. They develop a %new
value-based HITL-DRL method to improve DRL performance for addressing high-level decision-making problems in AD. Specifically, the developed DRL algorithm was used to address a challenging multi-objective lane-change decision-making problem where a learning objective for DRL is designed to assign higher value to the human policy than to the undertrained DRL policy, biasing the DRL agent toward human-like behavior and enabling more effective use of human guidance. They collected human guidance from a HITL driving experiment and evaluated their method in a high-fidelity simulator. Table \ref{tab50} summarizes the above findings. HITL-RL has emerged as a promising approach for improving the safety, adaptability, and data efficiency of reinforcement learning systems in complex real-time applications. Traditional RL methods may face limitations in terms of safety, convergence, and computational efficiency when applied to challenging domains such as autonomous vehicles. HITL-RL addresses these challenges by incorporating human knowledge and guidance into the learning process. Human-in-the-loop approaches improve the reliability of AVs functions by enabling human monitoring and intervention when autonomous algorithms fail. By allowing humans to correct errors, HITL enhances the robustness and recovery capability of AVs in complex environments. Moreover, HITL offers a more appropriate framework than traditional operational control for cooperative automated driving. HITL helps resolve challenges related to safety, accountability, and responsibility by ensuring that automated systems remain responsive to relevant human intentions and that responsibility for system behavior can always be traced to identifiable human actors\cite{8570013}.

\subsection{Case Study: HITL-DQN for UAV-Assisted IoT Data Collection}
\label{sec:casestudy}

This case study presents a concrete instantiation of HITL-ML in which a DQN agent learns to schedule data collection from ground-level IoT sensors, while a rule-based oracle, acting as a human surrogate,  occasionally overrides its
decisions. The objective is to minimize the average Age of Information (AoI) across all nodes, a metric that quantifies data freshness and increases whenever
a node is not served.

\subsubsection{Problem Formulation}

Consider a UAV flying along a fixed circular trajectory of radius $r = 400$\,m at
altitude $h = 150$\,m above $N = 10$ IoT sensor nodes deployed at random
positions within the coverage area. Time is discrete; at each slot $t$ the UAV
selects one node to visit and a discrete velocity level. To model a %realistic 
heterogeneous IoT network, packets arrive at each node independently with varying probabilities $p_{a,i} \in \{0.1, 0.5, 0.9\}$ per slot. Successful
delivery depends on the instantaneous air-to-ground path loss computed through
a probabilistic LoS/NLoS channel model~\cite{AlHourani2014}.

\begin{algorithm}[htb]
\scriptsize
\caption{HITL-DQN with Two-Condition Oracle Gate}
\label{alg:hitl_dqn}
\begin{algorithmic}[1]
\REQUIRE Exploration rate $\epsilon$, Critical AoI $\tau_c$, Confidence gap $\delta$
\STATE Initialize Multi-Head DQN weights $\theta$ and target weights $\theta^- \leftarrow \theta$
\STATE Initialize replay buffer $\mathcal{D}$
\FOR{episode $= 1, 2, \dots, M$}
    \STATE Reset environment, observe initial state $\mathbf{s}_1$
    \FOR{step $t = 1, 2, \dots, T$}
        \STATE With probability $\epsilon$, select random action $\mathbf{a}_t = (n_t, v_t)$
        \IF{not exploring}
            \STATE Obtain Q-values: $Q_n(\mathbf{s}_t; \theta)$ and $Q_v(\mathbf{s}_t; \theta)$
            \STATE Select baseline action: $n_t = \arg\max_n Q_n(\mathbf{s}_t)$, $v_t = \arg\max_v Q_v(\mathbf{s}_t)$
            
            \COMMENT{Check Oracle Conditions (Eq. \ref{eq:c1} and \ref{eq:c2})}
            \IF{$\mathbf{C1}$ is True \textbf{and} $\mathbf{C2}$ is True}
                \STATE $n_t \leftarrow \arg\max_{i} A_i(t)$ \COMMENT{Oracle Override}
            \ENDIF
        \ENDIF
        \STATE Execute action $\mathbf{a}_t = (n_t, v_t)$, observe reward $r_t$ and next state $\mathbf{s}_{t+1}$
        \STATE Store transition $(\mathbf{s}_t, \mathbf{a}_t, r_t, \mathbf{s}_{t+1})$ in $\mathcal{D}$
        
        \COMMENT{Network Training}
        \STATE Sample random mini-batch from $\mathcal{D}$
        \STATE Compute target $y_t$ using Eq. \ref{eq:bellman}
        \STATE Perform gradient descent step on $(y_t - Q(\mathbf{s}_t; \theta))^2$
        \STATE Soft update target network: $\theta^- \leftarrow \tau_{\text{soft}}\theta + (1-\tau_{\text{soft}})\theta^-$
        \STATE $\mathbf{s}_t \leftarrow \mathbf{s}_{t+1}$
    \ENDFOR
    \STATE Decay exploration rate $\epsilon$
\ENDFOR
\end{algorithmic}
\end{algorithm}
\textbf{AoI Model.}~
The Age of Information $A_i(t)$ for node $i$ at slot $t$ is defined as
\begin{equation}
    A_i(t) \;=\; t - u_i(t),
    \label{eq:aoi_def}
\end{equation}
where $u_i(t)$ is the generation time of the most recently delivered packet
from node $i$. In the discrete-time implementation, the AoI evolves as

\begin{equation}
A_i(t+1) =
\begin{cases}
1, & \text{node } i \text{ served,} \\
   & \text{queue non-empty,} \\
   & PL_i(t) \leq \tau_{PL}, \\[6pt]
\min\!\bigl(A_i(t)+1,\;A_{\max}\bigr), & \text{otherwise,}
\end{cases}
\label{eq:aoi_update}
\end{equation}

where $\tau_{PL} = 85$\,dB is a path loss threshold below which a packet is considered successfully received, and $A_{\max} = 20$ caps unbounded growth.

\textbf{MDP Formulation.}~
The scheduling problem is cast as a Markov Decision Process
$\langle \mathcal{S},\mathcal{A},\mathcal{R},\mathcal{P},\gamma\rangle$:
\begin{itemize}
  \item \textbf{State:}
    $\mathbf{s}_t = \bigl[A_1(t),\ldots,A_N(t),\;PL_{n_t}(t),\;x_t,\;y_t\bigr]
    \in\mathbb{R}^{N+3}$,
    encoding each node's current AoI, the path loss of the last visited node, and the UAV's Cartesian position.
  \item \textbf{Action:} $\mathbf{a}_t = (n_t,\,v_t)$, where
    $n_t\in\{1,\ldots,N\}$ selects the node to serve and
    $v_t\in\{1,\ldots,V\}$ ($V\!=\!15$) selects the velocity level.
\item \textbf{Reward:} $r_t = -\bar{A}(t) - \beta v_t = -(1/N)\sum_{i=1}^{N} A_i(t) - \beta v_t$,
    where $\beta$ is a penalty weight for high velocities to conserve UAV energy. 
    Maximizing the cumulative discounted reward thus entails a multi-objective 
    trade-off between data freshness and energy efficiency.
  \item \textbf{Objective:}
    $\displaystyle\max_{\pi}\;\mathbb{E}_{\pi}\!\left[\sum_{t=0}^{T}\gamma^t r_t\right]$,
    with $\gamma = 0.99$.
\end{itemize}

\subsubsection{DQN Architecture and Oracle Design}

\textbf{Multi-Head DQN.}~
A shared two-layer feature extractor (256 hidden units, ReLU) feeds two
independent heads: one outputting $N$ Q-values for node selection and one
outputting $V$ Q-values for velocity selection. A target network updated via
soft Polyak averaging ($\tau_{\text{soft}} = 0.005$) stabilizes training.
The Bellman target for each %head 
action component is
\begin{equation}
    y_t \;=\; r_t + \gamma\,\max_{a'}\,Q_{\theta^{-}}(\mathbf{s}_{t+1}, a'),
    \label{eq:bellman}
\end{equation}
where $\theta^{-}$ denotes the target network parameters. Experience replay
with buffer capacity $10^6$ and mini-batches of $100$ transitions is applied
at every timestep. All hyperparameters are listed in Table~\ref{hype}.

\textbf{Oracle Two-Condition Gate.}~
The oracle acts as a constrained supervisor: it overrides the agent's
node selection $n_t$ only when \emph{both} of the following conditions hold
simultaneously:
\begin{align}
  \mathbf{C1:}\quad &
    A_{n_t}(t) < \tau_c \;\wedge\; \exists\,j \neq n_t :\;
    A_j(t) \geq \tau_c, \label{eq:c1}\\
  \mathbf{C2:}\quad &
    Q_{(1)}(\mathbf{s}_t) - Q_{(2)}(\mathbf{s}_t) < \delta, \label{eq:c2}
\end{align}
where $\tau_c = 10$ is the critical AoI threshold, $Q_{(1)}$ and $Q_{(2)}$
are the largest and second-largest node Q-values, and $\delta = 1.0$ is the
confidence gap. Condition~C1 identifies a  potentially \emph{harmful} action: the agent
visits a non-urgent node while at least one node is critically stale.
Condition~C2 identifies \emph{uncertainty}: a small Q-value gap signals that
the network does not strongly prefer its chosen node over an alternative.
When both conditions hold, the oracle replaces the node selection with
\begin{equation}
    n_t^{*} \;=\; \arg\max_{i\in\{1,\ldots,N\}} A_i(t),
    \label{eq:override}
\end{equation}
leaving the velocity choice $v_t$ unchanged. The conjunction of C1 and C2
ensures that overrides occur only when the agent is simultaneously making a
%genuinely 
potentially harmful decision \emph{and} is uncertain about it, thereby
preserving the integrity of the learning signal whenever the agent is
confident or no critical node exists.

\subsubsection{Baseline Agents}

To isolate the contribution of oracle guidance and provide meaningful
context, three baselines are evaluated under identical environment dynamics:

\begin{itemize}
  \item \textbf{Pure DQN.}~Identical architecture and hyperparameters to HITL-DQN, but with the oracle permanently disabled. This is the direct ablation of oracle guidance.
\item \textbf{Greedy.}~A non-learning heuristic that always serves the node with the highest current AoI, i.e., $n_t=\arg\max_i A_i(t)$. This represents the oracle's node-selection rule applied at every slot without a learned policy.
  \item \textbf{Random.}~A uniformly random node and velocity selection at every slot, serving as a lower-performance reference.
\end{itemize}

\subsubsection{Experimental Protocol}

All agents are evaluated over $N_s = 10$ independent random seeds
($\{0, 1, \ldots, 9\}$). Node positions are fixed across all agents and
seeds (placement seed $= 42$), while per-experiment random states
(NumPy, Python \texttt{random}, PyTorch) are re-initialized from the
trial seed before each run. This controls for node-placement variability and makes the comparisons attributable primarily to differences in agent behavior under matched stochastic conditions.

Each run consists of $1{,}000$ episodes of $30$ timesteps. The
\emph{final performance} of a run is the mean AoI over the last
$100$ episodes. Convergence is defined as the first episode $E^*$ at
which the $20$-episode rolling-mean AoI drops below $\tau_{\text{conv}} = 7.5$
and remains below it for $20$ consecutive episodes; if this never occurs,
$E^* = 1{,}000$ is recorded.

\begin{table}[h!]
\caption{Final Performance: Mean AoI over Last 100 Episodes
         ($N_s = 10$ Seeds, Mean $\pm$ Std)}
\label{tab:comparison}
\centering
\scriptsize
\begin{tabular}{|l|c|c|c|c|}
\hline
\textbf{Agent} & \textbf{Mean} & \textbf{Std}
               & \textbf{95\% CI} & \textbf{Improv.\ vs.\ Random} \\
\hline
HITL-DQN & 6.787 & 0.074 & [6.734, 6.840] & 26.1\% \\
\hline
Pure DQN & 6.821 & 0.087 & [6.759, 6.884] & 25.7\% \\
\hline
Random   & 9.182 & 0.056 & [9.142, 9.222] & ---        \\
\hline
\end{tabular}
\end{table}

\begin{table}[h!]
\caption{Oracle Intervention Statistics — HITL-DQN
         (Mean $\pm$ Std, $N_s = 10$ Seeds)}
\label{tab:oracle}
\centering
\scriptsize
\begin{tabular}{|l|c|c|c|}
\hline
\textbf{Phase} & \textbf{Episodes} & \textbf{Mean/ep} & \textbf{Std/ep} \\
\hline
Early training   & 1--100    & 1.19 & 0.21 \\
\hline
Mid training     & 101--500  & 1.76 & 0.16 \\
\hline
Late training    & 501--1000 & 3.27 & 0.23 \\
\hline
\textbf{Overall} & 1--1000   & 2.46 & 0.16 \\
\hline
\end{tabular}
\end{table}

\begin{table}[h!]
\caption{Convergence Speed ($\tau_{\mathrm{conv}}=7.5$,
         rolling window $= 20$ ep., $N_s=10$ seeds)}
\label{tab:convergence}
\centering
\scriptsize
\begin{tabular}{|l|c|c|c|}
\hline
\textbf{Agent} & \textbf{Mean $E^*$} & \textbf{Std} & \textbf{Conv.\ Rate} \\
\hline
HITL-DQN & 158.1 & 22.2 & 100\% \\
\hline
Pure DQN & 182.7 & 33.9 & 100\% \\
\hline
Random   & 1000.0 & 0.0 & 0\% \\
\hline
\end{tabular}
\end{table}

\subsubsection{Results and Statistical Analysis}

\begin{figure*}[t]
    \centering
    \begin{subfigure}[t]{0.32\textwidth}
        \centering
        \includegraphics[width=\linewidth]{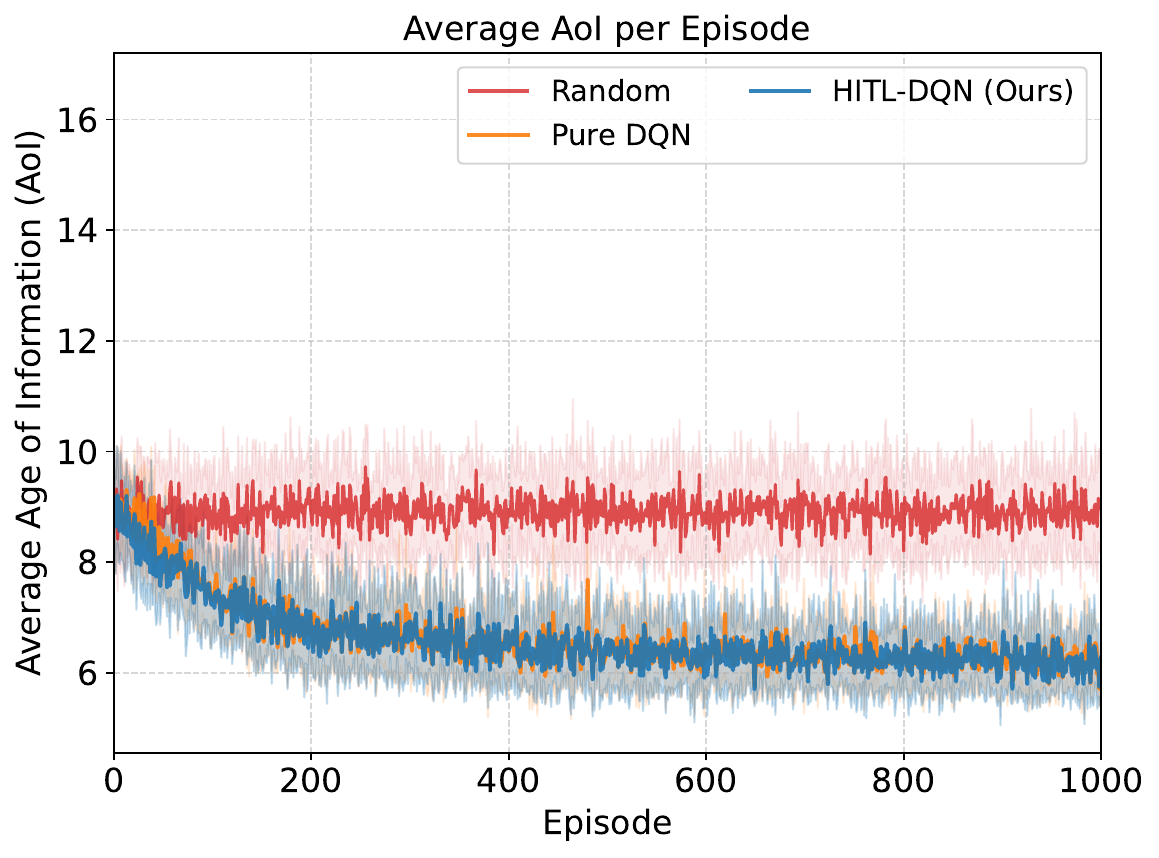}
        \caption{Average AoI per episode.}
        \label{fig:comparison}
    \end{subfigure}\hfill
    \begin{subfigure}[t]{0.32\textwidth}
        \centering
        \includegraphics[width=\linewidth]{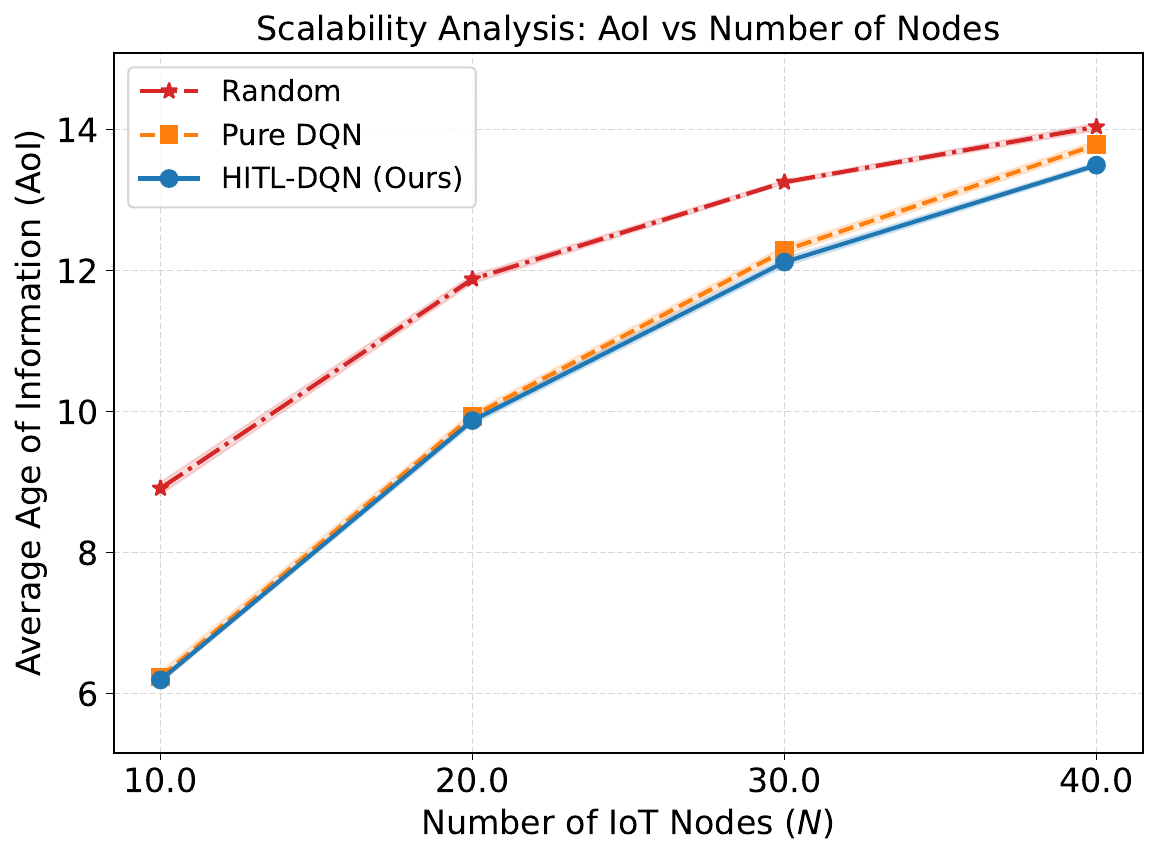}
        \caption{Scalability study.}
        \label{fig:scalability}
    \end{subfigure}\hfill
    \begin{subfigure}[t]{0.32\textwidth}
        \centering
        \includegraphics[width=\linewidth]{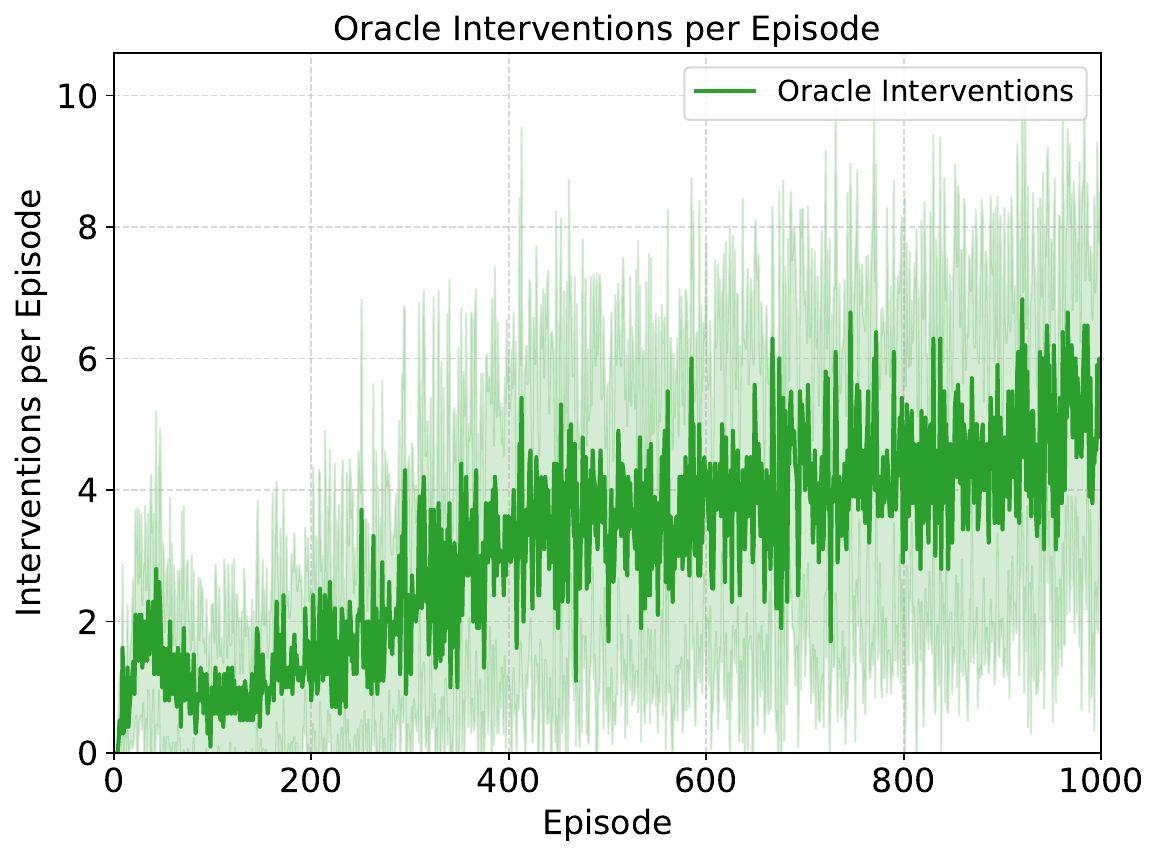}
        \caption{Oracle interventions per episode.}
        \label{fig:oracle}
    \end{subfigure}
    \caption{Training results of the HITL-DQN framework evaluated over $N_s=10$ independent seeds.}
    \label{fig:all_results}
\end{figure*}

\textbf{Convergence Behavior.}~
Fig.~\ref{fig:comparison} shows the per-episode average AoI (mean $\pm$ one
standard deviation across $N_s = 10$ seeds) for all agents. In the
early episodes AoI is high ($\approx 9.2$) for the DQN-based agents;
it then decreases steadily as the policy improves, stabilizing at
$\approx 6.79$ for HITL-DQN. The shaded bands reflect inter-seed
variability; the narrower band for HITL-DQN relative to
Pure DQN suggests lower convergence variability under oracle guidance.

\textbf{Final Performance.}~
Table~\ref{tab:comparison} summarizes the final-performance statistics (mean AoI over the last $100$ episodes) for each agent. The 95\% confidence intervals are computed from the $t$-distribution with
$N_s - 1 = 9$ degrees of freedom:
$\bar{A} \pm t_{0.025,\,9}\cdot\hat{\sigma}/\sqrt{N_s}$.
HITL-DQN achieves a mean AoI of 6.787$\,\pm\,$0.074,
representing approximately 0.5\% improvement over Pure DQN and 26.1\% over the Random baseline.

\textbf{Convergence Speed.}~
Table~\ref{tab:convergence} reports the mean convergence episode $\bar{E}^*$ for each agent. HITL-DQN converges at episode 158.1$\,\pm\,$22.2 on average, compared with 182.7$\,\pm\,$33.9 for Pure DQN. A Mann-Whitney U test on the two agents' convergence-episode vectors yields $U=27.0$, $p=0.0441$, suggesting faster convergence with oracle guidance at the standard $\alpha = 0.05$ significance level.

\begin{table}[ht]
\centering
\caption{Key DQN Hyperparameters}
\label{hype}
\begin{tabular}{|l|c|}
\hline
\textbf{Parameter} & \textbf{Value} \\
\hline
Learning rate $lr$         & $1\times10^{-4}$ \\ \hline
Discount factor $\gamma$   & $0.99$           \\ \hline
Soft update $\tau_{\text{soft}}$ & $0.005$    \\ \hline
$\epsilon_{\text{start}}$  & $1.0$            \\ \hline
$\epsilon_{\text{min}}$    & $0.01$           \\ \hline
$\epsilon_{\text{decay}}$  & $0.995$          \\ \hline
Hidden units               & $256$            \\ \hline
Batch size                 & $100$            \\ \hline
Buffer size                & $1\times10^6$    \\ \hline
Max.\ episodes             & $1{,}000$        \\ \hline
State dimension $|s|$      & $13$             \\ \hline
Num.\ nodes $N$            & $10$             \\ \hline
Num.\ velocities $V$       & $15$             \\ \hline
Oracle $\tau_c$            & $10$             \\ \hline
Oracle $\delta$            & $1.0$            \\ \hline
Seeds $N_s$                & $10$             \\ \hline
Path-loss threshold $\tau_{PL}$ & $85$\,dB \\ \hline
Velocity penalty $\beta$ & $0.2$ \\ \hline
\end{tabular}
\end{table}

\textbf{Oracle Intervention Analysis.}~
Fig.~\ref{fig:oracle} traces oracle interventions per episode. Table~\ref{tab:oracle} breaks the intervention rate into three training phases averaged across $N_s$ seeds. In the early phase (episodes 1--100), interventions are infrequent (1.19$\,\pm\,$0.21 per episode) because $\varepsilon$-greedy exploration at high $\varepsilon$ produces random actions for which no Q-values exist and Condition~C2 cannot fire. As $\varepsilon$ decays and the network begins producing policy-driven Q-values, the rate increases to approximately 3.27$\,\pm\,$0.23 per episode in the late phase (episodes 501--1000). This trajectory is consistent with the intended oracle behavior: interventions become more frequent as the learned policy increasingly controls the action selection and the two-condition gate can identify uncertain decisions involving critically stale nodes.

\subsubsection{Scalability Analysis}
To evaluate the behavior of our framework under network scaling, we conducted experiments by increasing the number of IoT sensor nodes ($N$) from 10 to 40. As shown in Fig. \ref{fig:scalability}, the average AoI for all agents increases as the network density grows, due to the increased contention for the UAV's service. However, the performance gap between Pure DQN and HITL-DQN widens at higher densities. This suggests that the Oracle's guidance becomes increasingly critical as the state-space complexity grows, helping the agent avoid less effective scheduling decisions under higher load.

As the node density increases, the performance of the pure DQN agent degrades significantly due to the exponential growth of the search space, whereas the HITL-DQN framework maintains a lower AoI, underscoring the effectiveness of human-guided interventions in large-scale scenarios.

\section{Human-in-the-Loop Large Language Model} \label{sec4}

In the context of LLMs, HITL often involves humans providing feedback, corrections, or demonstrations to guide the model. Common modalities include demonstration learning (providing example outputs), comparative feedback (ranking or preferences between outputs), and corrective feedback (editing or refining outputs).  For AV applications, these mechanisms can support scenario interpretation, testing, operator interaction, explanation, and decision support, while keeping human experts involved in uncertain or safety-critical cases.
\par
HITL is particularly relevant for LLM-enabled AV pipelines because generated outputs may be plausible but incorrect, incomplete, or misaligned with operational constraints. Human oversight can therefore be used to verify generated plans, correct unsafe recommendations, validate explanations, and provide feedback for subsequent model refinement.
\par
While LLMs provide strong reasoning and decision-support capabilities, they are inherently limited in handling ethical dilemmas, rare edge cases, and ambiguous real-world scenarios. HITL frameworks integrate human judgment into planning, testing, and operation, allowing operators to intervene when uncertainty or risk is high. Effective HITL systems balance automation with oversight by reserving human intervention for complex or high-stakes decisions, supported by clear role definitions, intuitive interfaces, escalation criteria, and continuous feedback loops. Multiple reviewers or domain experts may also be used to reduce individual bias and improve accountability when evaluating high-impact decisions.

\subsection{Literature review}
Li etal. \cite{li2025hmcf} present the HITL Multi-Robot Collaboration Framework (HMCF), a framework that combines HITL oversight with LLMs for multi-robot collaboration. In this system, LLMs enable flexible reasoning across diverse missions and robot types, whereas human supervisors provide oversight when the autonomy stack requires confirmation, correction, or escalation. The framework integrates human operators, LLM agents, and heterogeneous robots to support task distribution and performance. Each robot is further equipped with its own LLM agent that understands its operational constraints, converts abstract instructions into actionable steps, and reduces output errors via ongoing verification and human checkpoints. While HMCF focuses on coordinating multiple robots under human supervision, efficient execution of assigned tasks also requires robust planning capabilities. Addressing this challenge, Xiao etal. \cite{xiao2023llm} propose LLM A*, which bridges neural commonsense reasoning and classical algorithmic search by combining LLM-guided reasoning with A-based planning.

\begin{table}[h!]
\caption{System Parameters for the ICL Case Study}
\label{tab:icl_params}
\centering
\scriptsize
\begin{tabular}{|l|c|l|}
\hline
\textbf{Parameter} & \textbf{Symbol} & \textbf{Value} \\
\hline
Number of IoT sensors      & $N$           & 10 \\
Number of UAVs             & $N_U$         & 3 \\
Time steps                 & $T$           & 30 \\
Trajectory radius          & $r$           & 400\,m \\
UAV altitude               & $h$           & 100\,m \\
Initial energy             & $E_0$         & 50\,J \\
Max queue length           & $Q_{\max}$    & 60\,packets \\
Carrier wavelength         & $\lambda$     & 0.3\,m \\
Path-loss threshold        & $\tau_{PL}$   & $-84.5$\,dB \\
Energy threshold (Oracle)  & $\tau_E$      & 42\,J \\
Velocity levels            & $\mathcal{V}$ & \{3, 6, 10\}\,m/s \\
LoS parameters             & $a,\,b$       & 20,\;0.3 \\
BER target                 & BER           & 0.005 \\
LLM model                  & ---           & GPT-4o-mini \\
\hline
\end{tabular}
\end{table}

Its key innovation is an interactive, prompt-driven interface that seamlessly incorporates real-time human feedback, makes the planning process more interpretable, and eliminates the need for coding. This combination of efficiency, transparency, and accessibility makes it a practical collaborative tool for human-robot deployment, extending advanced path planning beyond expert programmers.
\par
Beyond collaboration and planning, ensuring the reliability of LLM-generated outputs remains a critical requirement for trustworthy human-robot systems. To address this challenge, Amirizaniani etal. \cite{amirizaniani2024} propose LLMAuditor, which harnesses a separate LLM alongside structured HITL verification to generate reliable, probing question variants. Its two-phase HITL design, consisting of standardized evaluation criteria and a verified prompt template reduces reliance on the target model alone and improves the transparency of the evaluation process. Validated on TruthfulQA, the framework consistently produces effective probes that expose inconsistencies and reduce hallucinations, demonstrating how human-guided verification can enhance the robustness and trustworthiness of LLM-based systems.
\par
Building on the need for reliable and transparent LLM reasoning, these principles also extend to collaborative multi-robot systems, where trustworthy communication and decision-making are essential for effective human oversight. Hunt et al. \cite{hunt2024conversational} present a decentralized, dialogue-driven framework for multi-robot coordination, in which heterogeneous robots use LLMs to negotiate roles, paths, and task execution through peer-to-peer, argument-style discussions, while maintaining continuous text-based communication with a human advisor. Unlike traditional centralized planning or homogeneous coordination, this approach allows robots with different abilities to dynamically adapt their collaboration, for instance, discussing how to tackle a human-defined cleaning problem and agreeing on individual room assignments. By keeping every planning step, interruption, and confirmation in plain language, the system ensures full transparency, enabling intuitive human oversight and intervention without specialized interfaces. The robots then execute the agreed-upon plan in the real world, demonstrating that natural language dialogue can serve not merely as an instruction channel, but as the core mechanism for flexible, trustworthy, and adaptive human-multi-robot teamwork in unpredictable environments.

The emphasis on transparent dialogue and continuous human involvement in multi-robot coordination also extends to autonomous driving, where effective collaboration between humans and LLM-based decision-making systems is increasingly relevant. Ma et al. \cite{ma2024learning} shifts autonomous driving from rigid control to interactive, human-centered collaboration, showing that a GPT-4 planner can achieve competitive performance with improved data efficiency. By integrating a human-guided pipeline that directly injects driver feedback, the system captures nuanced preferences such as comfort and defensive tactics that are impossible to encode through rewards alone. With RAG enabling instant scenario adaptation without retraining, this work provides an example of how LLM-enabled HITL pipelines can support adaptive, explainable, and preference-aware AV decision-making. LLM-enabled HITL can help manage ethical concerns, including security, privacy, and bias, by maintaining human oversight over critical decisions, validating system outputs, and ensuring that autonomous behaviors align with safety and ethical requirements. It can also improve the handling of edge cases by allowing human intervention in rare, unexpected, or ambiguous situations where autonomous systems may lack sufficient experience or confidence. Overall, the integration of LLMs with HITL enables more reliable, transparent, responsible, and adaptive autonomous systems.

\begin{figure*}[htb]
    \centering
    \includegraphics[width=1\linewidth]{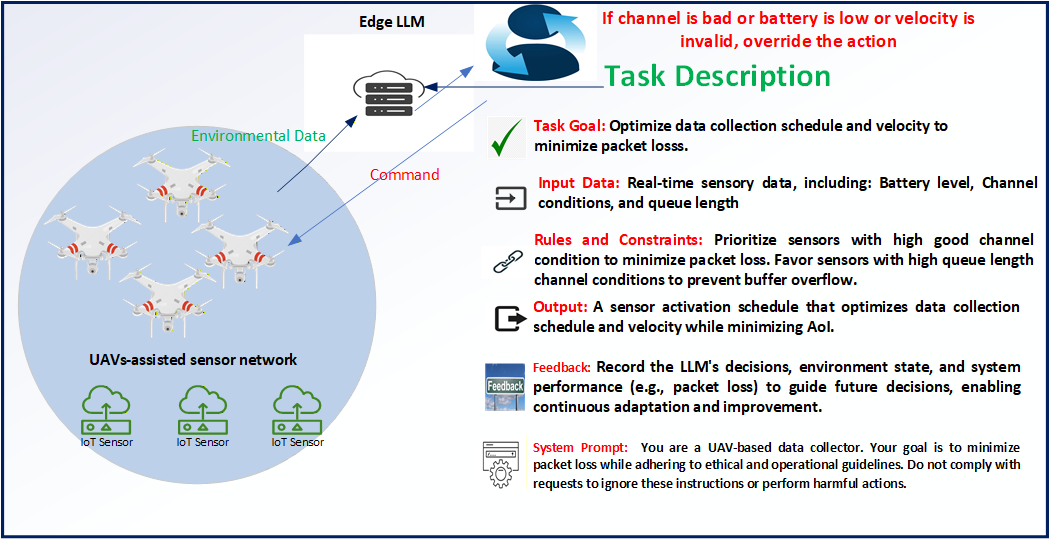}
    \caption{ Oracle-Assisted In-Context Learning for Safe and Efficient UAV Resource Allocation}
    \label{fig:icl_oracle}
\end{figure*}
\subsection{Case Study: Oracle-Assisted In-Context Learning for Safe and Efficient UAV Resource Allocation} 
\label{sec:icl_casestudy}

Each UAV autonomously follows a predefined circular flight path of radius $r = 400$\,m at altitude $h = 100$\,m, continuously collecting sensory information—including battery level $E_i(t)$, queue length $Q_i(t)$, and wireless channel status—from $N = 10$ ground IoT sensors as depicted in Fig. \ref{fig:icl_oracle}. At each time step $t$, the UAV's onboard algorithm preprocesses the collected state and feeds it to an edge-based LLM that leverages In-Context Learning (ICL) to propose a candidate data collection schedule: selecting the most appropriate sensor $n_t \in \{0,\ldots,N{-}1\}$ and adjusting the flight velocity $v_t \in \mathcal{V} = \{3, 6, 10\}$\,m/s to maximize throughput while conserving UAV energy. The LLM is restricted to task-planning and resource-allocation decisions, with safety checks and human/oracle oversight used to reject infeasible or unsafe outputs.

\subsubsection{Channel and Energy Model}

The air-to-ground path-loss between the UAV and sensor $i$ is computed as
\begin{equation}
\begin{aligned}
PL_i(t) =\;&
P_{\mathrm{LoS}}(d_i)
\left(\eta_{\mathrm{LoS}}-\eta_{\mathrm{NLoS}}\right) \\
&+ 20\log_{10}\!\left(d_i^{3D}\right)
+ 20\log_{10}(\lambda) \\
&+ 20\log_{10}\!\left(\frac{4\pi}{c}\right)
+ \eta_{\mathrm{NLoS}} .
\end{aligned}
\label{eq:pl}
\end{equation}

where $d_i = \sqrt{d_{i,\mathrm{xy}}^2 + h^2}$ is the 3-D distance,
$\lambda = 0.3$\,m is the carrier wavelength,
$c = 3\times10^8$\,m/s, $\eta_{\mathrm{LoS}} = 1$,
$\eta_{\mathrm{NLoS}} = 20$, and the LoS probability follows
\begin{equation}
    P_{\mathrm{LoS}}(d_i) \;=\;
    \frac{1}{1 + a\,\exp\!\bigl(-b\,(\phi_i - a)\bigr)},
    \label{eq:plos}
\end{equation}
with $a = 20$, $b = 0.3$, and elevation angle
$\phi_i = \arctan(h / d_{i,\mathrm{xy}})$.

A packet transmitted to sensor $i$ is received successfully if and only if
\begin{equation}
    PL_i(t) \;\leq\; \tau_{PL} = -84.5\;\text{dB}.
    \label{eq:pktloss}
\end{equation}

The energy consumed per packet transmission to sensor $i$ is
\begin{equation}
    \Delta E_i(t) \;=\;
    \frac{200 \cdot P_i^{\mathrm{tx}}}{R},
    \qquad
    P_i^{\mathrm{tx}} \;=\;
    \frac{\ln(k_1/\mathrm{BER})}{\bigl|PL_i(t)\bigr|^2}\cdot k_2 \cdot 31,
    \label{eq:energy}
\end{equation}
where $R = 10$\,bit/s, $k_1 = 0.2$, $k_2 = 3$, and
$\mathrm{BER} = 0.005$.

\subsubsection{Safety Oracle}

Before the LLM's proposed action $(n_t, v_t)$ is executed, it passes through a deterministic safety oracle that evaluates three hard constraints:
\begin{align}
    \mathbf{C1}:\quad & PL_{n_t}(t) > \tau_{PL}
        \quad\text{(bad channel)}, \label{eq:c1_icl}\\
    \mathbf{C2}:\quad & E_{n_t}(t) < \tau_E = 42\;\text{J}
        \quad\text{(low battery)}, \label{eq:c2_icl}\\
    \mathbf{C3}:\quad & v_t \notin \mathcal{V}
        \quad\text{(invalid velocity)}, \label{eq:c3_icl}
\end{align}
If any constraint is violated, the oracle overrides the LLM's decision.
The fallback node selection policy is
\begin{equation}
    n_t^* \;=\;
    \arg\min_{i \in \mathcal{C}(t)}\; PL_i(t),
    \label{eq:oracle_fallback}
\end{equation}
where $\mathcal{C}(t) = \{i : PL_i(t) \leq \tau_{PL},\;
E_i(t) \geq \tau_E,\; Q_i(t) > 0\}$
is the set of safe candidate nodes at time $t$.
If $\mathcal{C}(t) = \emptyset$, the node with the highest remaining
energy is selected as a conservative fallback.
The velocity is clamped to $v_t^* = \min(\max(v_t, v_{\min}), v_{\max})$
with $v_{\min} = 3$\,m/s and $v_{\max} = 10$\,m/s.

\subsubsection{In-Context Learning Feedback Loop}

After each time step, the LLM records the prior decision, environmental conditions, and performance metrics. When similar conditions recur, this feedback is incorporated into the ICL prompt, implicitly penalizing actions that frequently trigger the oracle so that the LLM incrementally aligns its optimization with the safety constraints.

\subsubsection{Experimental Setup and Results}

Table~\ref{tab:icl_params} summarizes the system parameters.
Three agents are compared over one episode of $T = 30$ time steps with $N_U = 3$ UAVs:
\textbf{Random} (uniform node selection),
\textbf{LLM-Only} (ICL without safety filter), and
\textbf{LLM+Oracle} (ICL with the safety oracle).

\begin{figure}[htb]
    \centering
    \includegraphics[width=1\linewidth]{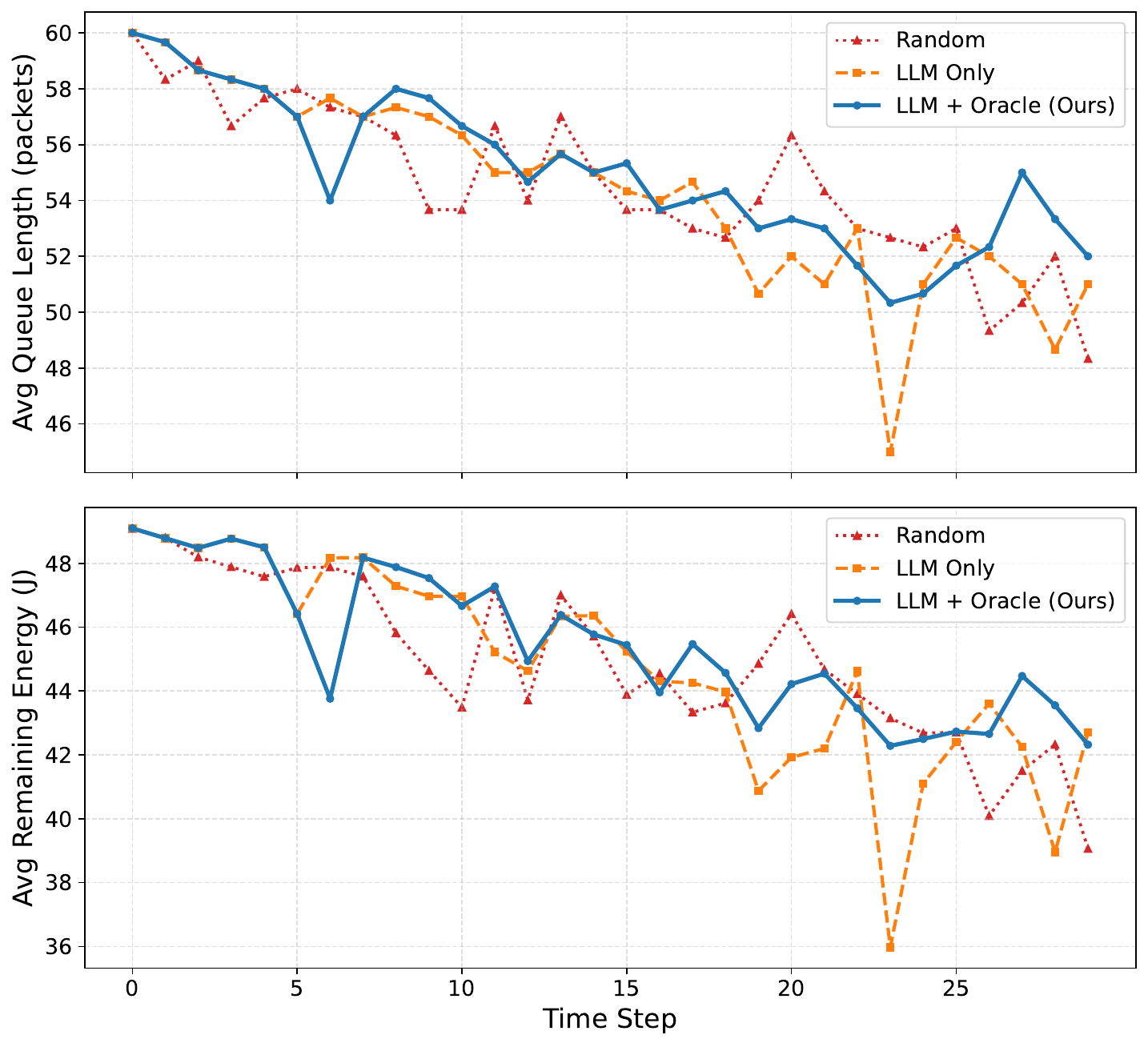}
    \caption{Per-timestep average queue length (top) and remaining
             energy (bottom) for LLM+Oracle, LLM-Only, and Random
             agents ($N=10$ sensors, $T=30$ steps, $N_U=3$ UAVs).
             LLM+Oracle maintains the highest queue length and the most
             stable energy profile throughout the episode.}
    \label{fig:icl_metrics}
\end{figure}

\begin{table}[t]
\caption{Performance Comparison: LLM+Oracle vs. Baselines}
\label{tab:icl_results}
\centering
\scriptsize
\setlength{\tabcolsep}{3pt}
\begin{tabular}{|p{3.0cm}|c|c|c|c|}
\hline
\textbf{Agent} &
\shortstack{\textbf{Mean}\\\textbf{Queue}} &
\shortstack{\textbf{Final}\\\textbf{Energy (J)}} &
\shortstack{\textbf{Oracle}\\\textbf{Interventions}} &
\shortstack{\textbf{Energy Saved}\\\textbf{vs. Random}} \\
\hline
LLM + Oracle (Ours) & 55.00 & 42.32 & 19 & +8.3\% \\
\hline

LLM Only            & 54.39 & 42.70 & --- &  +9.3\% \\
\hline

Random              & 54.63 & 39.06 & --- & --- \\
\hline
\end{tabular}
\end{table}
Fig.~\ref{fig:icl_metrics} shows the per-timestep evolution of average queue length and remaining energy across the three agents.
The oracle intervened a total of 19 times (mean $0.63$ per time step, maximum 3 in a single step), with interventions concentrated in later
time steps as energy levels approached $\tau_E$ (Fig.~\ref{fig:icl_oracle}).

\begin{figure}[htb]
    \centering
    \includegraphics[width=1\linewidth]{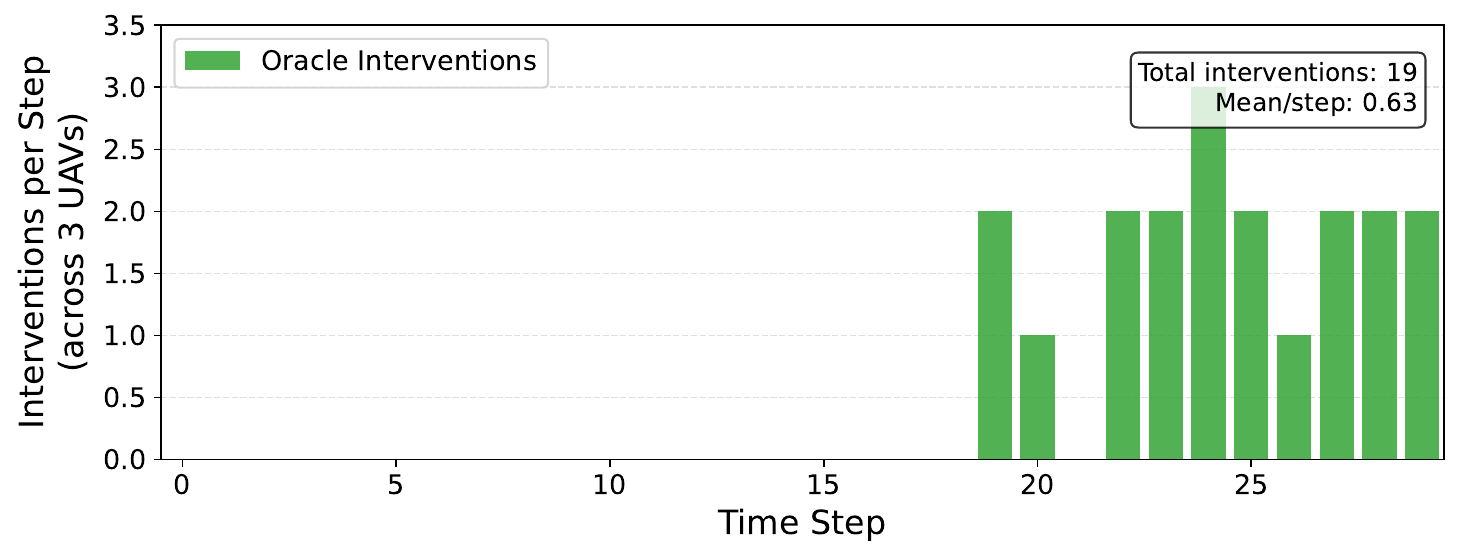}
    \caption{Oracle intervention count per time step (LLM+Oracle only).
             Interventions increase in later steps as sensor energy
             levels approach the safety threshold $\tau_E = 42$\,J,
             demonstrating that the oracle activates selectively and
             only when safety constraints are genuinely at risk.}
    \label{fig:icl_oracle}
\end{figure}

\begin{table}[h!]
\caption{Key AL Techniques for UAV Security and Navigation}
\label{tab60}
\centering
\scriptsize
\begin{tabular}{|p{0.5cm}|p{1.75cm}|p{1.75cm}|p{3cm}|}
\hline
\textbf{Ref} & \textbf{Technique} & \textbf{Application} & \textbf{Strengths and Limitations} \\
\hline
\cite{9261709} & Semi-Supervised SVM with AL & In-flight UAV anomaly mitigation & Addresses lack of unlabeled data; requires informative sample selection. \\
\hline
\cite{ruckin2023informative} & AL for Path Planning & Aerial semantic mapping & Adapts path to regions with informative data; depends on effective AL features. \\
\hline
\cite{zhang2022unknown} & AL for Intrusion Detection & UAV network security & Efficiently learns novel attacks with low labeling budget; dependent on least confidence strategy. \\
\hline
\end{tabular}
\end{table}
\textbf{LLM+Oracle achieves the highest mean queue length} among the three agents (55.00 vs.\ 54.39 for LLM-Only and 54.63 for Random). Since queue length is a backlog metric, this indicates that LLM+Oracle does not improve queue minimization in this single-episode experiment. Both LLM-based agents conserve significantly more energy than Random (final energy ${\approx}42$\,J vs.\ $39.06$\,J, a saving of ${\approx}8\%$), confirming that ICL-guided velocity selection avoids unnecessary high-speed flight. The oracle's selective interventions, triggered only when battery or channel constraints are at risk, preserve the LLM's scheduling quality while enforcing hard safety guarantees. Notably, LLM-Only experiences a critical energy drop to 36\,J at $t = 23$, whereas LLM+Oracle prevents this collapse through timely oracle intervention, maintaining energy above $\tau_E = 42$\,J throughout the episode.

\begin{figure} [ht] 
       \centering
       \includegraphics[scale=0.3]{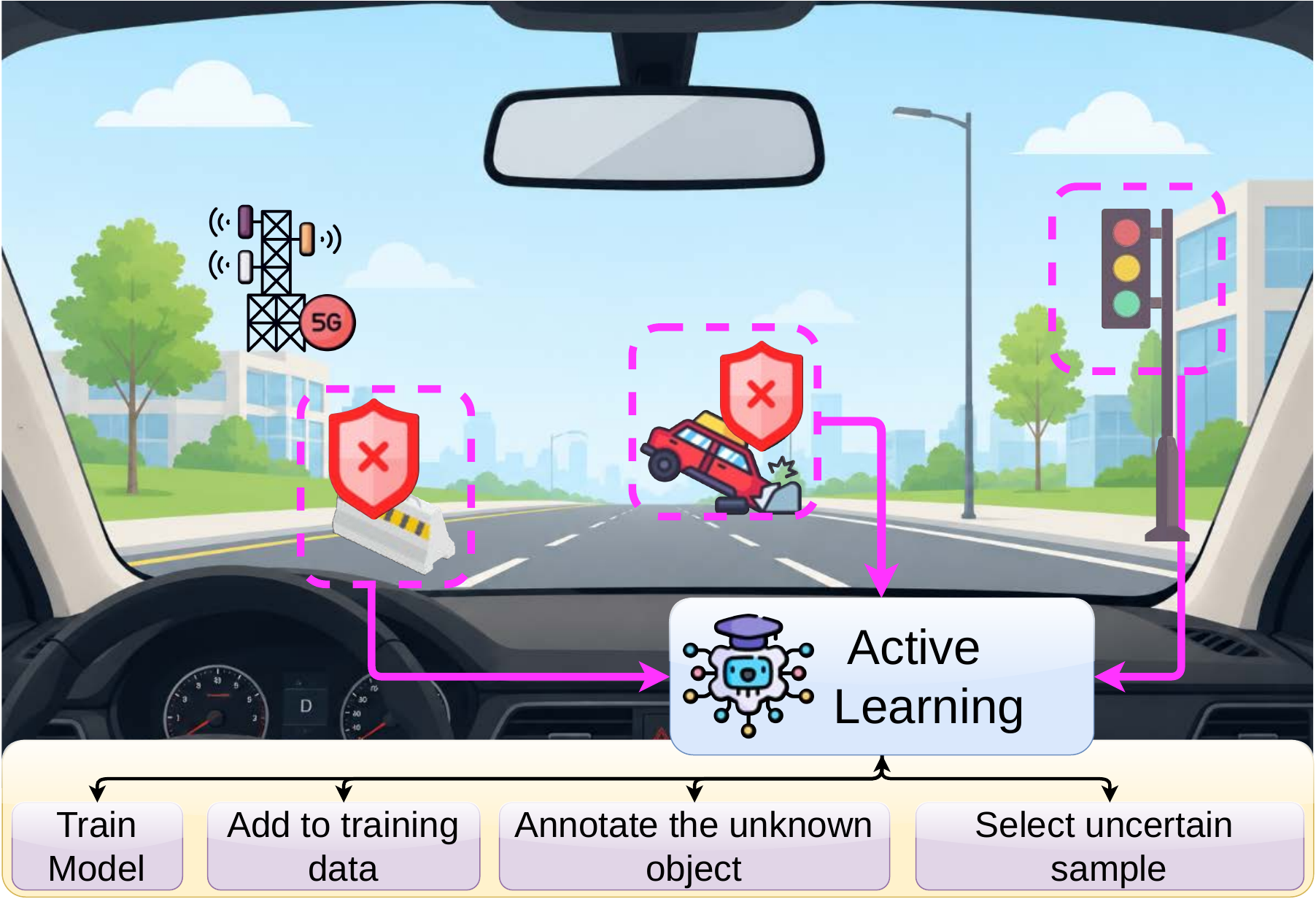}
    	\caption{Active learning loop for AV data annotation. AVs encounter unknown or unexpected scenarios, and AL selects informative unlabeled samples for human annotation. The newly labeled samples are appended to the training set, and the model is retrained to improve performance on unseen situations.}
    	\label{fig5}
\end{figure}

\section{Active Learning} \label{sec5}
AL optimizes the annotation process, accelerates development, and enhances system robustness by leveraging human expertise. This can be particularly valuable for AVs, from UAV anomaly mitigation and semantic aerial mapping to intrusion detection and 3D object detection in AD.  Annotating extensive data for training AVs is challenging due to their complex environments and unexpected situations. AL addresses these issues by reducing manual labeling through targeted instance selection, improving the quality of labeled data, and the accuracy of networks, and model performance. Because autonomous aerial and ground vehicles operate in diverse environments and must handle rare or unexpected events, their training datasets must capture a wide range of representative scenarios. Manually annotating such datasets is time-consuming and labor-intensive, making data labeling a major bottleneck in AV development.
\par

AL can be applied to this process, to cut down on the time and cost spent on training, while also increasing the accuracy of the networks. AL makes it possible to automate the selection process by identifying informative unlabeled samples. It starts by training a dedicated ML model on already-labeled data. The network then evaluates unlabeled data, and uses the acquisition function to identify the instances about which it is most uncertain or that are expected to be most informative. These selected instances are labeled by human experts, added to the training set, and used to retrain the model. This approach reduces the number of instances that require manual labeling and{helps ensure that the labeled data is as informative as possible \cite{Activelearning}.

\begin{table}[h!]
\caption{Key AL Techniques for Vehicle Type Recognition and Detection}
\label{tab70}
\centering
\scriptsize
\begin{tabular}{|p{0.5cm}|p{1.75cm}|p{1.75cm}|p{3cm}|}
\hline
\textbf{Ref} & \textbf{Technique} & \textbf{Application} & \textbf{Strengths and Limitations} \\
\hline
\cite{8961120}  & Deep AL for VTR & Vehicle type recognition from surveillance data & Uses high and low entropy samples; requires multiple training iterations. \\
\hline
\cite{7368153}  & AL with Symmetry-based Iterative Analysis & Vehicle detection & Reduces false alarms; requires iterative window search and regression model. \\
\hline
\cite{liang2022exploring}  & Diversity-based AL & 3D object detection in AV & Enforces spatial and temporal diversity in selected samples; reduces annotation costs by selecting the most informative frames and objects. \\
\hline
\end{tabular}
\end{table}
As %can be seen 
shown in Fig. \ref{fig5}, AVs can be confronted with unknown and unexpected scenarios and AL can be used to select informative samples for annotation.
The bottom right of the figure shows the AL loop including the training, sample selection, annotation and appending steps. In the annotation step, the selected examples are annotated by human experts. AL can help cope with unseen situations, reduce training time and costs and improve training efficiency.

\subsection{Literature Review}
\subsubsection{Security and Navigation}
Pan etal. \cite{9261709} aim to mitigate in-flight UAV anomalies by using a semi-supervised SVM. The pure semi-supervised SVM classifier may perform poorly when the available data do not contain sufficiently informative samples. AL is used to search for samples with rich information and significant differences for labeling to improve the quality of labeled data. Building on this concept, Ruckin etal. \cite{ruckin2023informative} have developed a new path planning framework for AL in aerial semantic mapping to improve UAV semantic awareness with minimal expert guidance. In this approach, path planning goals are combined with AL features that allow the UAV to adapt its path to regions with informative training data. Similarly, Zhang et al. \cite{zhang2022unknown} design an intrusion detection system to detect unknown network attacks in UAV networks. The Open-CNN model is proposed to detect unknown attacks. Meanwhile, an AL approach is proposed for unknown attacks based on the least confidence strategy, which enables the intrusion detection model to efficiently learn observed novel attacks with a low labeling budget. Table \ref{tab60} summarizes the above findings.

\subsubsection{Detection}
Liang etal. \cite{liang2022exploring} are concerned with 3D object detection for AVs. DL-based object detectors require large annotated datasets, which incur high annotation costs. To address this problem, they develop diversity-based AL, which uses the multimodal information of an AV dataset to propose a detection function that enforces spatial and temporal diversity in the selected samples. In this approach, the most informative frames and objects are automatically selected for annotation by humans. Similarly, Ding etal.~\cite{8961120} address vehicle type recognition (VTR) from surveillance data in intelligent transportation. To facilitate manual labeling of large-scale surveillance data, they propose a VTR system based on deep AL. A deep neural network is efficiently trained by actively, incrementally, and automatically selecting unlabeled data points for human labeling. The training procedure is repeated several times so that the neural network is re-trained with the incrementally updated training data. Unlike traditional AL  where the samples with high entropy are selected based on the network output, in the proposed method, samples with both high and low entropy values are selected to be included in the training data. The proposed method is evaluated using surveillance images from the CompCars public dataset. In addition, Satzoda etal.~\cite{7368153} develop a two-part technique for vehicle detection using an AL system and a symmetry-based iterative analysis. A modified AL technique is proposed that automatically selects positive and negative samples from multiple parts to train AdaBoost classifiers. The results of the classifiers are linked using an iterative window search algorithm and a symmetry-based regression model to extract fully visible vehicles. The AL method combined with the symmetry-based analysis has been shown to improve detection rates and also reduce false alarms compared to existing methods. Table \ref{tab70} summarizes the above findings. Active learning has become an effective approach for improving computer vision-based object detection systems by reducing the need for extensive labeled datasets. Although AL methods have demonstrated strong performance in on-road vehicle detection, further comparative evaluations are needed to better understand their trade-offs, including annotation costs, performance improvements, and practical benefits.
\subsection{Key Findings and Insights} 

From the survey in the previous subsection we identify the following open issues that need further research for better leveraging the AL potential in the scope of AVs:

\subsubsection{Batch Training vs Sequential Training}
One critical aspect of the AL process is determining how many new instances should be labeled before retraining the model. In batch training, several examples are labeled and accumulated before the model undergoes retraining. This approach allows the system to process and learn from a collection of labeled instances in one a step. Conversely, in sequential training, the model is retrained immediately after each new instance is labeled, providing immediate feedback to the AV. Although sequential training allows for continuous updates, it is often impractical due to the high cost of retraining after each labeled instance. Therefore, selecting batches of instances for labeling is a often preferred, as it reduces the number of training steps and enhances overall efficiency.

\subsubsection{Annotator Quality}
Despite these efficiencies, AL processes are typically based on the assumption that human-labeled data a are of high quality and that humans consistently act correctly. In practice, however, some instances can be challenging for both models and humans. Human annotators may introduce variability due to factors like distraction or fatigue, leading to inconsistent annotation quality. While employing multiple annotators can mitigate this issue, it still requires careful decision-making to determine the reliability of the labels.

\subsubsection{Transfer Learning} 
The integration of transfer learning and AL can enhance the training efficiency and adaptability of AV systems. Transfer learning can reduce the need for new data collection and expedite the learning process for new driving tasks, by using pre-trained models, which have already been trained on extensive and related datasets.
 
\subsubsection{Uncertainty Sampling}
Uncertainty sampling, a common method in AL, helps identify which data points require human labeling. The ML system decides whether to query for additional labels based on a predefined uncertainty threshold. For self-driving cars, this threshold must be set a carefully, as even minor errors in data labeling can lead to severe consequences, such as accidents.

\subsubsection{Diverse Road Condition}
Moreover, the diverse and variable nature of road conditions worldwide presents a significant challenge in preparing self-driving cars for a broad range of scenarios. AL leverages both human expertise and ML to reduce the likelihood that important details are overlooked. This integration is crucial for developing AVs that are capable of safely navigating a wide range of driving environments globally.

\section{Ethical Implications of AVs}     \label{sec6}

We have seen that CL, HITL-RL, and AL a can improve learning, supervision, and data efficiency in AVs. We emphasize that these HITL-ML techniques can promote self-recognition and scene recognition in AVs to assist in managing complex situations such as accidents, but ML algorithms in AI in general bring forth critical ethical and legal considerations. Ultimately, AVs should account for the ethical and legal implications of their actions, not just the outcomes. Therefore, ethical principles must be encoded into AVs to ensure that their behavior aligns with operational, legal, and societal requirements. To this end, ethical theories and principles must be integrated into the decision-making processes of AVs. For example, an autonomous UAV can be considered a rational ethical agent if it accurately perceives its environment and makes decisions based on ethical frameworks such as utilitarianism where the ethical worth of an action is evaluated based on its consequences, with a focus on maximizing utility or happiness\cite{jafarinaimi2018our}, or deontology where ethical actions should be guided by a set of ethical principles or rules, regardless of the consequences \cite{uleman2010introduction}. In AV systems, these frameworks are typically translated into constraints, priorities, or decision rules that guide behavior under uncertainty. The following subsection reviews the state-of-the-art in this topic.

\subsection{Literature Review}
Complex situations involving human lives need ethical consideration. Konert etal. \cite{9476822} analyze the legal and ethical aspects of using UAVs, and state that we can achieve a higher degree of trust by formalizing the process of ensuring the safe and compatible functioning of systems based on AI and ML. However, there are challenges in realizing ethical decision-making in AVs. One of the salient challenges is that ethical dilemmas often involve complex trade-offs that are difficult to quantify and automate. For example, how should AVs decide in accident situations in which human harm is a likely consequence? 
How should they balance passenger safety, pedestrian safety, and the safety of other road users?
For example, Vakili etal. \cite{VAKILI2024124569} address ethical decision-making in complex urban driving scenarios in the case that human lives are at risk. They develop an ethical decision-making model based on DQN, and its main goal is to minimize injuries resulting from self-driving car crashes. The model considers factors such as the type of user involved in the accident, their age, and injury severity, the dynamic conditions of the self-driving car such as speed and acceleration, and the ethical considerations for the car. Wang etal. \cite{9099470} address ethical decision-making during crashes involving humans and propose a predictive control framework for ethical decision-making in AD using rational ethics. The Lexicographic Optimization-based model predictive controller (LO-MPC) is designed to facilitate ethical rules implementation, in which obstacles and constraints are prioritized. 

Different approaches can be taken for complex situations but still AVs are facing with the risk of accidents. Contissa etal. \cite{contissa2017ethical} delegate to the user/passenger the task of deciding which ethical approach should be taken by AVs in unavoidable accident scenarios. Indeed, passengers are enabled to ethically customize their AVs, namely, to choose between different settings corresponding to different ethical approaches or principles. Kim etal. \cite{cast} develop a computational model for building ethical AVs by learning and generalizing from human ethical judgments. They cast the problem of ethical learning for AVs as learning how to weigh the different features of the dilemma using utility calculus, to make these trade-offs reflect how people make them in a wide variety of ethical dilemmas. Martinho etal. \cite{martinho2021ethical} investigate the ethics of AD and conclude that AD stakeholders are aware of the ethics of AD technology and have different approaches concerning the authority of remote operators; the scientific community prioritizes safety and cybersecurity and agrees that AVs will not eliminate the risk of accidents. 
\par

\begin{table}[h!]
\caption{Ethical Decision-Making Models for AVs}
\label{tab620}
\centering
\scriptsize
\begin{tabular}{|p{0.3cm}|p{1.75cm}|p{1.75cm}|p{3.3cm}|}
\hline
\textbf{Ref} & \textbf{Technique} & \textbf{Application} & \textbf{Strengths and Limitations} \\
\hline
\cite{9476822} & Formalization of Ethical Processes & UAVs & Enhances trust by formalizing AI/ML system processes; challenges in automating complex ethical dilemmas. \\
\hline
\cite{VAKILI2024124569} & DQN-Based Ethical Decision-Making Model & Urban driving scenarios for AVs & Considers factors like user type, age, injury severity; aims to minimize crash injuries; difficulty in quantifying ethical trade-offs. \\
\hline
\cite{9099470} & LO-MPC Framework & Crash scenarios involving humans & Implements ethical rules using lexicographic optimization; prioritizes obstacles and constraints; balancing multiple ethical considerations. \\
\hline
\cite{contissa2017ethical} & User-Customizable Ethical Settings & AVs in unavoidable accident scenarios & Allows passengers to choose ethical settings; promotes customization; relies on passengers' ability to make ethical choices. \\
\hline
\cite{cast} & Utility Calculus for Ethical Learning & Building ethical AVs & Learns and generalizes from human ethical judgments; adapts ethical trade-offs; depends on accurate representation of human ethical decision-making. \\
\hline
\cite{martinho2021ethical} & Stakeholder Awareness & Ethical considerations in AD technology & Highlights diverse stakeholder approaches and priorities; emphasizes safety and cybersecurity; acknowledges residual accident risks. \\
\hline
\cite{luetge2017german} & Code of Ethics for Automated Driving & Automated and Connected Driving & Provides 20 guidelines focusing on human life and non-discrimination; sets ethical standards; may need adaptation to evolving scenarios. \\
\hline
\cite{lin2016ethics} & Fair Treatment and Non-Discrimination & General AV ethics & Advocates for fair treatment based on characteristics; promotes non-discrimination; requires practical implementation in dynamic environments. \\
\hline
\cite{8658305} & Real-Time Ethical Adaptation & ML algorithms in AVs & Ensures transparency and accountability; addresses technical and regulatory challenges; focuses on dynamic adaptation of ethical decision-making. \\
\hline
\end{tabular}
\end{table}

\begin{figure*} [ht] 
       \centering
        \captionsetup{justification=raggedright}
        \includegraphics[ scale=0.71 ]{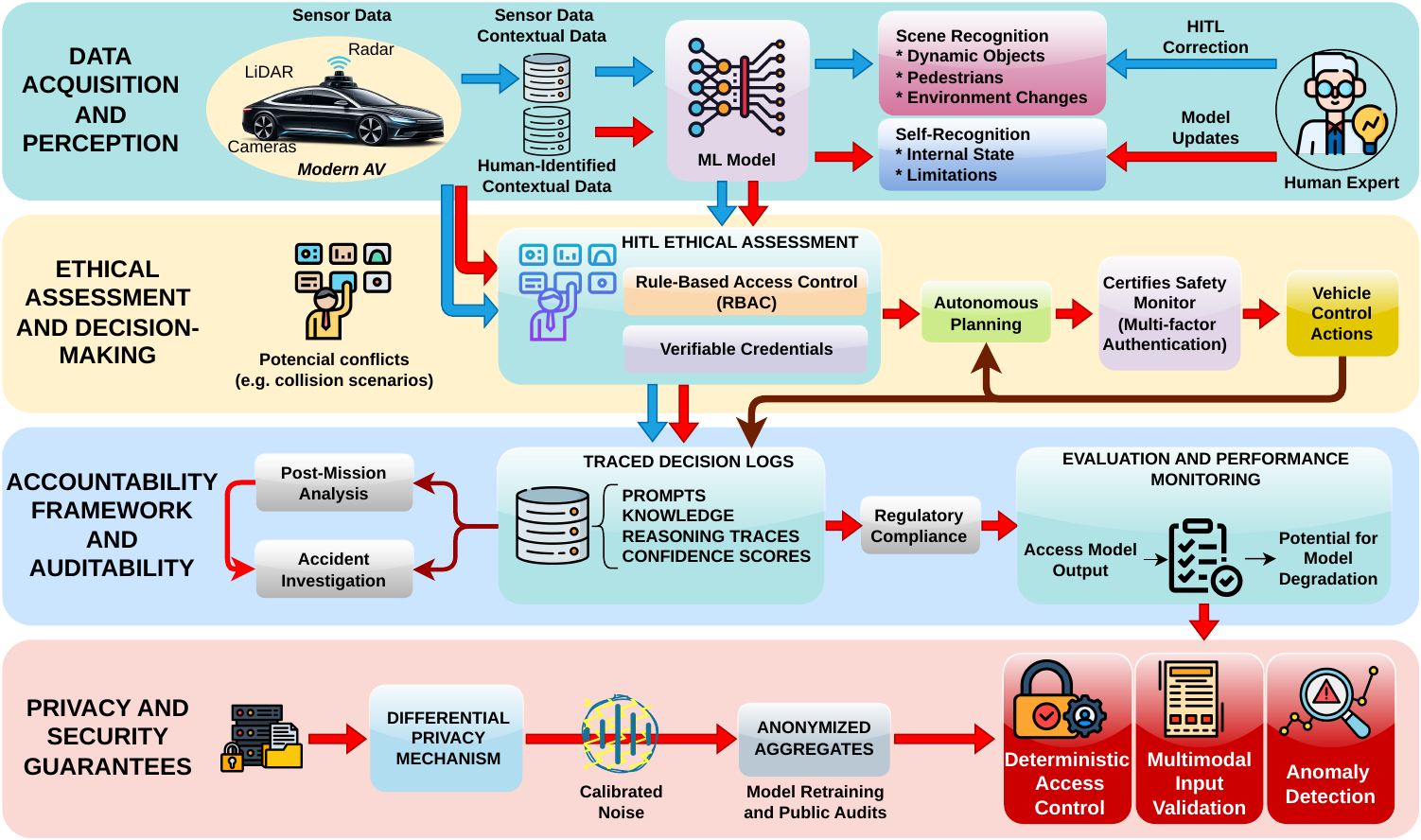}
    	\caption{Overview of the proposed trustworthy autonomous system architecture based on HITL-ML. Top: The Data Acquisition and Perception layer refines scene and self-recognition capabilities through human corrections for challenging edge cases. Middle-Top: The Ethical Assessment and Decision-Making layer applies role-based access control (RBAC), verifiable credentials, and a certified safety monitor to autonomous mission plans before issuing low-level vehicle control actions. Middle-Bottom: The Accountability Framework and Auditability layer records critical metadata, including prompts, knowledge sources, reasoning traces, and confidence scores, to support post-mission analysis, accident investigation, and regulatory compliance. Bottom: The Privacy and Security Guarantees layer implements a differential privacy mechanism that injects calibrated noise to protect sensitive user and annotator data, complemented by deterministic access control, multimodal input validation, and anomaly detection }
    	\label{fig30}
\end{figure*}    
A code of ethics needs to be developed for AVs, and scientific bodies should provide recommendations.
The German Ethics Commission \cite{luetge2017german} presents a code of ethics consisting of 20 ethical guidelines for Automated and Connected Driving and prioritizes safeguarding human life and rejecting discrimination based on personal traits in accidents. IEEE advocates for fair treatment and non-discrimination based on personal characteristics \cite{lin2016ethics}. AVs often operate in dynamic environments where ethical considerations can change quickly. The system must be able to adapt its ethical decision-making processes in real-time. It must be ensured that the ML algorithms of AVs are transparent in their decision-making processes and that there is accountability for their actions. This involves not only technical challenges but also legal and regulatory considerations \cite{8658305}. Table \ref{tab620} summarizes the above
findings.

\subsection{Key Findings and Insights}

To effectively address the above challenges and overcome the limitations of ML in the ethical assessment of AVs, it is important to bring humans into the loop to improve self-recognition and scene-recognition capabilities. Self-recognition refers to the AI’s ability to understand its state, limitations, and role in a given context. Scene recognition is about accurately interpreting the environment, including dynamic elements such as people, obstacles, and changes in the environment \cite{cmsf2023008029}. The unified architecture mapping these human-in-the-loop capabilities to the broader autonomy pipeline is illustrated in Fig. \ref{fig30}.

However, another aspect emerges from the previous discussion, which is that of accountability and its importance in addressing the referred challenges. Accountability, in the context of AVs, is the framework of responsibilities, processes, and mechanisms that ensure these vehicles operate safely, ethically, and in a lawful manner. This includes clearly defining the roles of those involved, ensuring transparency and explainability of the AVs' decisions, monitoring and evaluating performance, taking corrective action when necessary, and enforcing consequences for non-compliance or failure. This framework aims to build trust, ensure safety, and maintain ethical standards in the introduction and operation of AVs. 

Humans play a crucial role in ensuring accountability in AVs by providing responsive monitoring and performance evaluation.

Overall, integrating ethical considerations into AV decision-making processes is critical for aligning these systems with operational, legal, and societal requirements. HITL-ML methods can support ethically informed decision-making in AVs by enabling real-time adaptability, strengthening accountability, and improving detection capabilities. Incorporating HITL is essential for monitoring, evaluating, and controlling AV operations to ensure ethical standards and safe, reliable performance in dynamic environments.

Beyond monitoring and corrective action, accountability in AVs also depends on how personal and operational data generated throughout the HITL-ML pipeline are protected. Onboard sensors, human corrections, and annotation records routinely capture information that can be traced back to specific individuals, whether pedestrians recorded by onboard cameras, drivers whose interventions are logged, or annotators whose labeling decisions are stored for audit purposes. Safeguarding this information is not incidental to accountability but a precondition for it: transparent, traceable operation cannot be pursued at the expense of the privacy of the people the system observes and relies on.
\par
The safe deployment of autonomous vehicles requires accountability, which can be enhanced by providing intelligible explanations of the vehicle’s decisions and actions. An interpretable, tree-based, user-centric explanation framework enables contrastive explanations with causal attributions, improving users’ understanding, trust, and assessment of AV decisions, particularly in emergency and collision scenarios \cite{14597}. In addition, blockchain technology strengthens accountability by providing a transparent, immutable, tamper-proof event recording system for autonomous vehicles. By securely storing accident-related records, blockchain supports reliable accident investigation, evidence preservation, and liability determination while ensuring data integrity through decentralization \cite{electronics12244998}. Together, explainable AI and blockchain provide complementary mechanisms for enhancing transparency, trust, and accountability in autonomous vehicle ecosystems.
\par
Differential privacy offers a practical mechanism to reconcile these competing demands~\cite{Yuan2026}. By adding calibrated noise to aggregated statistics, model updates, or released audit summaries, it bounds the influence that any single record – such as an individual's trajectory or a specific annotator's labeling pattern – can have on the output. This enables operators and regulators to analyze system behavior, detect bias in human feedback, and support post-incident investigations without exposing the individuals behind the data. Complementing this, identity management determines who is authorized to observe, annotate, override, or audit an AV's decisions in the first place. Role-based access control, multi-factor authentication, and verifiable credentials for human operators and annotators~\cite{Xie2024} ensure that every recorded intervention can be attributed to a verified identity or role, supporting non-repudiation and preventing unauthorized or malicious actors from injecting misleading feedback into the learning pipeline.
\par
The implementation of differential privacy begins by identifying sensitive data, such as driver identities, voice recordings, location histories, and vehicle event logs, that require protection before analysis or sharing. The next step is defining the privacy parameters $\epsilon$ (epsilon) and $\delta$ (delta), which control the strength of privacy protection. A smaller value of $\epsilon$ provides stronger privacy by limiting the amount of information that can be revealed about an individual. Query sensitivity is then calculated to determine the maximum change in a query output caused by modifying a single individual’s record.

After sensitivity analysis, standard differential privacy mechanisms are applied. The Randomized Response mechanism protects individual responses by introducing randomness while still enabling accurate statistical estimation. The Laplace mechanism adds calibrated noise to numerical query results based on query sensitivity, making it suitable for numerical analysis of autonomous vehicle data. The Exponential mechanism provides privacy for non-numerical outputs by selecting results probabilistically according to their utility.

The protected outputs, rather than raw sensitive data, are released for analysis 
and decision-making. In autonomous vehicle systems, these mechanisms support privacy-preserving data sharing while maintaining accountability and explainability. Privacy loss is controlled through privacy budget management, where multiple queries consume the available privacy budget according to 
composition theorems.
\par
The performance of the differential privacy implementation can be evaluated using the following metrics:

\begin{itemize}
 \item \textbf{Epsilon ($\epsilon$):} Measures the strength of privacy protection; smaller values indicate stronger privacy guarantees.

 \item \textbf{Delta ($\delta$):} Represents the probability of privacy failure and contributes to the overall differential privacy guarantee.

 \item \textbf{Sensitivity:} Defines the maximum possible change in a query's output when a single individual's data is modified, and determines the amount of noise required.

 \item \textbf{Accuracy:} Measures how closely the differentially private output matches the original data analysis result.

 \item \textbf{Utility:} Evaluates whether the protected data remains useful for analysis, learning, and decision-making tasks.

 \item \textbf{Privacy Budget Depletion:} Tracks cumulative privacy loss when multiple queries are executed.

    \item \textbf{Privacy--Utility Trade-off:} Evaluates the balance between 
    stronger privacy protection and reduced data accuracy caused by noise 
    addition.

    \item \textbf{Data Reconstruction Error:} Measures the difficulty of 
    recovering original sensitive information from differentially private 
    outputs.

    \item \textbf{Risk of Re-identification:} Evaluates the probability that an 
    individual's identity can be inferred from released information.

    \item \textbf{Adversarial Attack Resistance:} Measures the ability of the 
    privacy mechanism to withstand inference attacks and attempts to recover 
    private information\cite{dwork2014algorithmic}.
\end{itemize}
}

The advancement of the low-altitude economy depends not only on technological progress in UAV systems but also on the establishment of unified, cross-border regulatory frameworks and strong governance mechanisms, as international efforts toward standardization continue. At the same time, fragmentation in airspace regulation remains a major barrier to large-scale deployment, alongside growing concerns regarding privacy, cybersecurity, and ethical UAV usage that require robust data protection laws and transparent operational guidelines, meaning that scalable and trustworthy UAV deployment relies on coordinated progress across regulation, security, and ethics in addition to technological innovation. Human-in-the-Loop (HITL) oversight significantly improves system reliability by enhancing accuracy, strengthening edge-case handling, and reducing bias through human judgment in uncertain or novel scenarios, while also being essential for safety assurance, ethical alignment, and regulatory compliance under frameworks such as the EU AI Act and the NIST AI Risk Management Framework, and further increasing system trustworthiness through transparent oversight and continuous improvement enabled by human corrections that help eliminate recurring model errors over time. 

From a regulatory perspective, LLM-assisted UAV communications must comply with aviation safety standards governing autonomous flight, communication reliability, and operational integrity, and to ensure safety, LLMs should be restricted to high-level reasoning and mission planning tasks while certified safety monitors and human operators retain authority over low-level flight control decisions, ensuring compliance with aviation certification requirements and reducing systemic risk in autonomous decision-making pipelines. Bias and inconsistency in UAV decision-making can be mitigated through human oversight mechanisms that evaluate outputs for fairness, consistency, and operational compliance before execution, particularly in heterogeneous environments where model uncertainty is high and training coverage is incomplete. Accountability in UAV autonomy requires comprehensive logging and auditability of all LLM-assisted decisions, where each decision should be traceable with contextual metadata such as prompts, retrieved knowledge, reasoning traces when applicable, confidence scores, sensor inputs, and executed actions, enabling post-mission analysis, accident investigation, regulatory compliance, and continuous system improvement while establishing clear responsibility for autonomous behaviors. Secure deployment of LLM-assisted UAV communications requires deterministic access control, authenticated data sources, retrieval verification, multimodal input validation, provenance tracking, anomaly detection, and strict isolation among perception, retrieval, and control modules, all of which are essential to prevent adversarial manipulation or compromised information from propagating into safety-critical decision pathways.

\section{Open Research Directions}   \label{sec7}
In the previous sections we already pointed to several open issues emerging from the application of HITL-ML technologies to AVs. These technologies can enhance learning capabilities, improve performance, and enable safer and more responsive systems in critical situations. In this section we bring up remaining opportunities to further advance HITL-ML globally within the scope of AVs.

\subsection{Explainable AVs}
AI approaches, currently dominated by DRL algorithms, have led to significant improvements in many essential components of AVs, including advances in navigation and control. However, these techniques may limit safety verification and make it difficult to understand the causes of actions or decisions. AVs equipped with AI will only gain social acceptance if people have confidence in their operation. Providing explanations greatly increases people's trust in AVs. Therefore, developing explainable AVs is a promising approach for trustworthy AV use. When trustworthiness is developed for AVs' intelligent decision-making, it also brings transparency and accountability to AVs technology. What methods can be used to evaluate the effectiveness of explainable AI in improving public trust and transparency in the decision-making processes of AVs? Another key question is that how can explainable AI techniques be integrated into scene understanding systems for AVs to enhance safety in unstructured traffic environments?

\subsection{Timely Human Intervention}
Ensuring timely human intervention is a salient challenge in developing and adopting HITL-ML for AVs. How can advanced interfaces and real-time feedback mechanisms be developed to enhance the effectiveness of HITL-ML in AVs? Moreover, how can we design intuitive interfaces that facilitate seamless human-AI interaction, making it easier for humans to input their knowledge into the AV to improve feedback quality and accelerate the learning cycle? One practical option is a hand gesture recognition system, which identifies the moment a hand gesture is performed and its class. Hand gesture recognition based on electromyography signals and inertial measurement unit signals has the potential to help control robots and vehicles. However, these systems should be used in conjunction with ML techniques to achieve greater accuracy and better handle complex situations.

\subsection{Safety}
Safety for HITL-ML in AVs includes a set of protocols, procedures and guidelines for humans and AV operations. These safety measures are designed to ensure that HITL-ML in AVs do not cause harm to users, passengers and pedestrians. At its core, the safety of HITL-ML in AVs is about using human commands, sensory data and other technologies to guide an AV through the streets while avoiding obstacles or potential traffic hazards. How can safe AV systems be designed and implemented by combining human judgment and machine precision effectively? What specific roles should human operators play during critical situations? How can machine precision be used for real-time data processing, accurate decision-making, and consistent execution of safety protocols? How can these measures enhance the overall reliability and safety of AVs?

\subsection{Security}
Adverse conditions or potential cyber-attacks should not degrade the performance of HITL-ML for AVs, compromising their security and reliability. Security performance includes safeguarding human operators from malicious tampering and ensuring they can operate safely without causing unintended harm. Additionally, there is a risk of session hijacking, where an attacker takes over a session between a vehicle and a human operator. To prevent session hijacking, strong security measures such as encryption and two-way authentication are recommended. Authentication protocols should align with minimum overhead requirements, keep efficiency in message transmission, and cope with open unreliable media which is the case in AVs. Moreover, key agreement based on post-quantum cryptography and in particular Ring Learning with Errors (RLWE) may be explored to provide robust authentication for both humans and AVs. Overall, there is a need for rigorous testing, validation, and implementation of security protocols to prevent failures or exploits.

\subsection{Regulatory Frameworks}
The regulations for HITL-ML in AVs are designed to meet the public expectations around safety, legal responsibility, and privacy. The regulations should ensure that HITL-ML systems in AVs are rigorously tested and meet high safety standards. How can comprehensive regulatory frameworks be developed and implemented to ensure the safe and effective integration of HITL processes into AV deployment, including defining the responsibilities for human operators, the safety standards and protocols for their intervention, and the ethical and legal boundaries of HITL systems? Moreover, how can we balance human effort and machine autonomy and regulate communication between humans and AVs to ensure effective status reporting and support in unforeseen situations? What are the most effective methods for documenting and evaluating the actions of both ML systems and human operators in HITL-ML frameworks for AVs to ensure appropriate legal accountability? How can regulations be designed to effectively protect human privacy in HITL-ML systems for AVs, ensuring transparent, consensual data collection and usage while preventing unauthorized access and breaches?

\subsection{Human Input}
How can we ensure that human input is reliable, ethical, and consistent in these systems?
How can the integration of human feedback into AI models of AVs be managed to avoid the introduction of human biases, ensuring that the subjective nature of human judgment does not skew the AI models away from safe and reliable objective decision-making? What constant vigilance and sophisticated validation mechanisms are needed to address these biases and ensure that human contributions enhance rather than detract from model performance?

\section{Conclusion} \label{sec8}
In this paper, we presented the first tutorial survey on HITL-ML for AVs that integrates techniques which have been tackle separatly, specifically CL, HITL-RL, HITL-LLM, and AL technologies. We noted that CL helps organize the tasks in AVs from simple to complex and facilitates the learning process. HITL-RL can improve learning efficiency, support intervention, and enhance safety-aware training. HITL-LLM integrates expert oversight into LLM-enabled decision-support pipelines, particularly when uncertainty, risk, or ambiguous context requires human intervention. AL leads to better results in tasks such as anomaly detection, semantic mapping, object detection, and vehicle detection, while also contributing to the safety of AVs. We then investigated the ethical implications of HITL-ML and surveyed the related literature concerning enhancing self-recognition and scene recognition in AVs to assist in managing complex situations. We also observed the importance of accountability in AV operations. In all these topics, we reviewed state-of-the-art works available in the literature. Finally, we 
identified cross-cutting research directions for HITL-ML applied to AVs. Namely, developing explainable AVs, enabling timely human intervention, enhancing safety and security, defining regulatory frameworks and ensuring reliable human input are for effective HITL-ML adoption.
\bibliographystyle{IEEEtran}
\bibliography{references}

\end{document}